\newif\ificml
\icmltrue
\icmlfalse

\ificml %%%%%%%%%%%%%%%%%%%%%%%%%%% IF ICML
\documentclass{article}
\usepackage{microtype}
\usepackage{graphicx}
\usepackage{subfigure}
\usepackage{booktabs} % for professional tables

\usepackage{hyperref}

\usepackage{icml2025}

\usepackage{amsmath}
\usepackage{amssymb}
\usepackage{mathtools}
\usepackage{amsthm}
\usepackage[capitalize,noabbrev]{cleveref}

\theoremstyle{plain}
\newtheorem{theorem}{Theorem}[section]
\newtheorem{proposition}[theorem]{Proposition}

\theoremstyle{definition}
\newtheorem{definition}[theorem]{Definition}

\theoremstyle{remark}

\usepackage[textsize=tiny]{todonotes}

\icmltitlerunning{Submission and Formatting Instructions for ICML 2025}

\else %%%%%%%%%%%%%%%%%%%%%%%%%%% ELSE ICML
\documentclass[a4paper,11pt]{article}

\usepackage{amsmath}
\usepackage{amsfonts}
\usepackage{amssymb}         %additional math symbols (see amsguide.tex)
\usepackage{amsopn}
\usepackage{theorem}          %additional LaTeX2e package (see pctex\latex2e)
\usepackage{latexsym}
\usepackage{graphicx}        %graphic and .eps files
\usepackage{mathrsfs}		%calligraphic fonts
\usepackage{psfrag}
\usepackage{algorithmic}
\usepackage{algorithm}
\usepackage{color}
\usepackage{verbatim}
\usepackage[headings,twoside]{fancyhdr}
\usepackage{hyperref}
\usepackage{url}

\newtheorem{result}{\ }[section]
\theoremstyle{changebreak}                % (see LATEX2E\THEOREM.DTX)
\newtheorem{thm}[result]{Theorem}
\newtheorem{defn}[result]{Definition}

\newtheorem{prop}[result]{Proposition}

\newenvironment{proof}
 {{\sl Proof.}\hspace*{1 ex}}%
 {{\nopagebreak\hspace*{\fill}$\Box$\par\vspace{12pt}}}
\renewcommand{\vec}[1]{\underline{#1}}
\newcommand{\transpose}[1]{{#1}^{\top}}

\makeatletter
\DeclareRobustCommand{\cev}[1]{%
  {\mathpalette\do@cev{#1}}%
}
\newcommand{\do@cev}[2]{%
  \vbox{\offinterlineskip
    \sbox\z@{$\m@th#1 x$}%
    \ialign{##\cr
      \hidewidth\reflectbox{$\m@th#1\vec{}\mkern4mu$}\hidewidth\cr
      \noalign{\kern-\ht\z@}
      $\m@th#1#2$\cr
    }%
  }%
}
\makeatother

\fi %%%%%%%%%%%%%%%%%%%%%%%%%%% ENDIF ICML

\usepackage{enumerate}
\usepackage{xcolor}
\usepackage{mdframed}
\definecolor{offwhite}{RGB}{245,245,245}
\mdfsetup{
  skipabove=12pt, skipbelow=12pt,   % space around the block
  innertopmargin=10pt, innerbottommargin=10pt,
  innerleftmargin=10pt, innerrightmargin=10pt,  % padding inside
  roundcorner=0pt, middlelinewidth=0pt,         % no rounded corners / mid rule
  splittopskip=\baselineskip, splitbottomskip=0pt, % nicer page breaks
  splitbottomskip=4pt,   % << adds a little space *below* last line on page
  needspace=1\baselineskip                        % avoid starting with a stub line
}
\newmdenv[
  backgroundcolor=offwhite,
  linecolor=offwhite,      % hide frame by matching color
  leftmargin=0pt, rightmargin=0pt
]{greyblock}

\ificml %%%%%%%%%%%%%%%%%%%%%%%%%%% IF ICML

\begin{document}

\twocolumn[
\icmltitle{Mathematics with large language models as provers and verifiers}



\icmlsetsymbol{equal}{*}

\begin{icmlauthorlist}
\icmlauthor{Hieu {Le Duc}}{equal,where}
\icmlauthor{Leo Liberti}{equal,lix}
\end{icmlauthorlist}

\icmlaffiliation{where}{T\'el\'ecom SudParis, Institut Polytechnique de Paris, 91000 \'Evry-Courcouronnes, France}
\icmlaffiliation{lix}{CNRS LIX Ecole Polytechnique, Institut Polytechnique de Paris, 91128 Palaiseau, France}

\icmlcorrespondingauthor{Hieu {Le Duc}}{duc-hieu.le@telecom-sudparis.eu}
\icmlcorrespondingauthor{Leo Liberti}{leo.liberti@polytechnique.edu}

\icmlkeywords{LLM, mathematics, theorem proving}

\vskip 0.3in
]



\printAffiliationsAndNotice{}  % leave blank if no need to mention equal contribution

\else %%%%%%%%%%%%%%%%%%%%%%%%%%% ELSE ICML

\begin{document}

\thispagestyle{empty}
\begin{center} 

{\LARGE Mathematics with large language models as provers and verifiers}
\par \bigskip
{\sc Hieu Le Duc${}^{1}$, Leo Liberti${}^{2}$} 
\par \bigskip
\begin{minipage}{15cm}
\begin{flushleft}
{\small
\begin{itemize}
\item[${}^1$] {\it T\'el\'ecom SudParis, Institut Polytechnique de Paris, 91000 \'Evry-Courcouronnes, France} \\ Email:\url{duc-hieu.le@telecom-sudparis.eu}
\item[${}^2$] {\it LIX CNRS, \'Ecole Polytechnique, Institut Polytechnique de Paris, F-91128 Palaiseau, France} \\ Email:\url{liberti@lix.polytechnique.fr}
\end{itemize}
}
\end{flushleft}
\end{minipage}
\par \medskip \today
\end{center}
\par \bigskip

\fi %%%%%%%%%%%%%%%%%%%%%%%%%%% ENDIF ICML

\begin{center}
  \begin{minipage}{0.8\textwidth}
    \textbf{\color{red}This paper presents mathematical arguments obtained from a protocol based on LLMs from OpenAI. We are currently in the process of telling apart the correct from the wrong proofs produced by our protocol. One of the problems is that the formalization step, partly carried out by humans, takes much longer than expected. We are currently asking human experts to verify the proofs. We shall post further versions of this report as the verification work progresses. Meanwhile, for the moment, none of the ``proofs'' mentioned herein should be taken as correct.}
  \end{minipage}
\end{center}

\begin{abstract}
  During 2024 and 2025 the discussion about the theorem-proving capabilities of large language models started reporting interesting success stories, mostly to do with difficult exercises (such as problems from the International Mathematical Olympiad), but also with conjectures \cite{feldmanKarbasi} formulated for the purpose of verifying whether the artificial intelligence could prove it. In this paper we report a theorem proving feat achieved by ChatGPT by using a protocol involving different prover and verifier instances of the \textsf{gpt-5} model working collaboratively. To make sure that the produced proofs do not suffer from hallucinations, the final proof is formally verified by the \textsf{lean} proof assistant, and the conformance of premises and conclusion of the \textsf{lean} code is verified by a human. Our methodology is by no means complete or exact. It was nonetheless able to solve five out of six 2025 IMO problems, and close almost a third of the sixty-six number theory conjectures in \cite{cohenprimes}. 
\end{abstract}

\section{Introduction}
This paper describes a protocol, involving the OpenAI Application Programming Interface (API) with a minimal human interaction, which was able to solve five out of six problems of the International Mathematical Olympiads (IMO), close almost a third of sixty-six open conjectures on cyclic numbers proposed in \cite{cohenprimes}, as well as discover and prove several new theorems in a variety of fields of mathematics. The human intervention only arises to verify that a formal version of the generated theorem statement is consistent with its semi-formal natural language description. More precisely, we only ask the human to verify the conformance of all premises and the conclusion of the formal and the semi-formal versions of the proof\begingroup \renewcommand{\thefootnote}{\fnsymbol{footnote}}\footnote[1]{At the time of publishing this technical report we are still carrying out the human part of the protocol: no proof should not be trusted yet.}\endgroup.

Alan Turing conceived his computational model of the computer not only as a mechanical way of performing computations \cite{turing}, and not only as a precise definition of the Leibnitzian \textit{calculemus} dream, with roots in Renaissance steganographers, whose influence crept through the centuries to reach Hilbert's tenth problem and its vague ``finite process'' to solve polynomial equations in integers. In the last years of his life, Turing defined machine intelligence by means of a probabilistic game played between a (Turing) machine and a human exchanging messages in natural language \cite{mind}. If a human evaluator with access to the messages fails to reliably tell the machine apart from the human, the machine is deemed to be as intelligent as the human.

This test, initially called by Turing the ``imitation game'', but today simply known as the ``Turing test'', held a prominent position in Artificial Intelligence (AI) literature since its birth. In \cite{mind}, Turing forecasted that by the year 2000 people would have attuned to computers being ``intelligent'' (according to his definition thereof). Over the years, commentators and AI experts made all sorts of predictions about the time when the Turing test would be passed by a computer. Most of them predicted the latest possible times at which they could reasonably expect to be alive \cite{AIpred}.

The Turing test situation changed abruptly with the advent of Large Language Models (LLM), a Machine Learning (ML) paradigm consisting in complicated architectures of interconnected Artificial Neural Networks (ANN) of extremely large size, and trained on huge textual datasets to predict the next word of an answer to a question posed by a human in natural language. Since the mid-2020s, the Turing test was passed by ChatGPT \textsf{gpt-4.5} LLM \cite{chatgptTuringTest}. We remark that no theory of learning involving ANNs guarantees exact results: typically, we are more in the ``probably approximately correct'' (PAC) setting \cite{pacbook}: with respect to an unknown ground truth, the likelihood of producing more or less correct predictions from sufficiently large training sets is reasonably high. Moreover, it is well known that LLMs suffer from ``hallucinations'': wrong predictions involving mistakes that no humans would make \cite{llm_hallucination}. 

Rather independently of these developments, computer science devised computer programs to help humans prove \cite{vampire} and verify \cite{lean4} theorems by means of formal language constructs. Since checking is easier than solving, there are more verifiers than provers. Unlike the PAC learning theory mentioned above, all proof assistants (verifiers) and most theorem provers are based on formal systems (FS) \cite{fs}, and therefore give rise to exact and guaranteed results. 

The whole endeavour proposed herein, then, appears a bit paradoxical: we are employing an LLM without exactness guarantees, trained on natural language text and prone to hallucinations, in order to prove formal theorems. But consider the following: (i) ANNs have been devised by analogy with the neurons that humans have in the brain; (ii) nontrivial theorem proving has been an exclusively human activity until very recently; (iii) humans do not come with guarantees either, and may also hallucinate, or in any case get proofs wrong. So that paradox is only apparent.

The essential point is that proving theorems requires creativity, and intuition plays a prominent part in human creativity: analogies, metaphors, similitudes, even daring comparisons that short-circuit logical arguments are often used by humans in devising (or discovering?) proofs. Automated theorem provers, being mostly based on FS, heuristically explore the countably infinite digraph of formal sentences derived from given axioms by recursively applied inference rules: the validity of every sentence follows from its premises. Since this graph is infinite, the search for the path leading from the axioms to the conclusion of the desired theorem statement may be indefinitely long, and establishing unprovability yields a non-terminating search (this is exactly the ``Entscheidungsproblem'' that Turing settled in \cite{turing}). It seems that LLMs may provide useful shortcuts in proving theorems, which is an invaluable feature even at the cost of inexactness and errors. Human brains, errors notwithstanding, establish proofs or the unprovability of formal sentences by a different process, without guarantees but generally with better success rates than those achieved by FS-based computer programs. The most visible link between humans and LLMs being natural language, there is a remarkable interest in using natural language (and thus LLMs) in having computers prove theorems. 

The rest of this paper is organized as follows: Sect.~\ref{s:litrev} summarizes the relevant literature. In Sect.~\ref{s:protocol} we describe our protocol in depth, and analyse its strengths and weaknesses. Then follows the description of the implemented system in Sect.~\ref{s:impl}. The results are described in Sect.~\ref{s:results}. Sect.~\ref{s:concl} concludes the paper. 

\subsection{Contributions of this paper}
The main contribution of this paper is a conceptually simple protocol, involving the repeated interaction of different LLM-based computational agents and terminating with a final human verification, that proves theorems. This protocol solved five out of the six problems of the 2025 IMO, exactly like the specialized (and as far as we know undisclosed) LLMs built by Google DeepMind and OpenAI; it solved several theoretical conjectures in \cite{cohenprimes}; and it discovered and solved new conjectures in three fields of mathematics (see Sect.~\ref{s:resmode} and \ref{s:results:resmode})\footnote{None of these announced results is sure yet: the proofs have been produced in semi-formal natural language, but the verification process involves a mixture of human and automatic work, which takes longer than initially anticipated}. Obviously, the protocol does not guarantee that every conjecture will be proved or disproved: after a given amount of time it can give up empty-handed.

A smaller contribution is the minimization of the human interaction, which is limited to verifying the conformance of premises and conclusion between the formal and the natural language versions of the proof of interest. This is enough to guarantee correctness of the formal version of the proof (Sect.~\ref{s:correctness}) despite the possibility of hallucinations (Sect.~\ref{s:hallucinations}), under the assumption that the human does not make any mistakes. 

A more subtle contribution, which may not sound new to LLM experts but may surprise traditional computer scientists, is the idea that ``programming with an LLM'' involves rhetorical skills in giving natural language instructions to LLM-based agents to tell them how to behave. Much like the grammarians of the late middle ages who fought over the vocative of \textit{ego}, LLM programmers often resort to various imperative forms ``to be'': we tell our LLM agents ``you are a meticulous researcher'', ``be cautious'', ``be concise'', and even ``you are a senior research assistant who is finding advanced sources in the literature to tackle an open problem''. These \textit{prompts} are passed to LLM instances to prepare them for their work and define their output (see Sect.~\ref{s:prompts}). The new field of \textit{computational pragmatics} is arising from such needs \cite{llm_behavior,llm_prompt,llm_prompt_eng}, but most of the literature takes ``prompt engineering'' to mean different data structures (sequences, trees, tables) containing inter-dependent prompts and corresponding answers, all of which are used to provide additional training to the LLM at hand. The prompts we use in our paper are of the ``zero-shot'' type with no additional training \cite{liZeroPrompt,anamZeroPrompt}.

\section{A review of the relevant literature}
\label{s:litrev}
Contributions reporting the successful use of LLMs to prove theorems are extremely recent: so recent, in fact, that the corresponding articles are often in the form of unrefereed technical reports.

Such contributions describe either fully automated or semi-automated approaches. In fully automated ones, LLMs are asked to devise proofs, and then encode them as programs for proof assistant languages, which are then executed to either deny or accept the proof produced by the LLMs. Wrong proofs are either discarded or modified \cite{seedprover,lyra}. In semi-automated approaches there is at least one step of the protocol that is assigned to a human: at best this is a verification step \cite{feldmanKarbasi}, but sometimes humans also guide\footnote{One of us (LL) has once enlisted the help of ChatGPT to reconstruct a proof of a result he had seen in the past but had forgotten, and could not afford the time to reconstruct by himself: the creativity originated, through the semi-formal use of natural language, from both parties (human and machine), in a protocol involving ChatGPT as the prover and the human as the verifier. The final verification was embodied by the full human re-write of the proof in the context of the paper where the proof was needed. Moreover, Scott Aaronson reported on Sept.~27th, 2025 on his \href{https://scottaaronson.blog}{blog} that he and a co-author used ChatGPT to obtain the proof of a lemma needed in a larger scope; he believes that they would have been able to prove this lemma without ChatGPT  ``in a week or two'', and adds that the proof was obtained by a prover/verifier protocol where he was the verifier.} the LLM around invalid inferences \cite{funsearch}. 

The problem of LLM ``hallucinations'' is by now well known \cite{llm_hallucination}. Employing a fully automated process for proving theorems therefore exposes the proof to the risk of hallucinations. The interaction with a proof assistant such as \textsf{lean} \cite{lean4} decreases the risk of hallucinations, but it is not eliminated \cite{liu_safe,leanCopilot,autoformalization}.

Consider this scenario: \label{hallucinating_scenario} the LLM generates a proof and the corresponding proof assistant code, both of which may be subject to hallucinations. The hallucinated code may still be accepted as valid by the proof assistant as long as the premises lead to the conclusion by inference, but these premises may not correspond to those of the LLM-generated proof. While we agree with \cite{liu_safe,autoformalization} that such an occurrence is unlikely, we also think that a wrong proof of this kind would be extremely difficult to catch by humans exactly because it is known to have been checked by a proof assistant. Our process overcomes this difficulty by introducing a limited human interaction: a human verifies the conformance of all premises and conclusion of the proof between the proof assistant code and the LLM-generated proof\footnote{Humans also make mistakes when proving theorems, but at this stage of LLM technology development we should be happy if computers do not make more mistakes than we do.}.

The interaction of humans and AI agents in the context of proving theorems is not new. See for example the end of \cite[\S 2]{leanCopilot} for a few cases. In \cite{leanCopilot}, the interaction between the human and the AI agent is based on the agent being a ``co-pilot'': the human leads the investigation but can ask the AI agents for suggestions. In our case the human is limited to a verification task to be carried out only once, at the end of the computer-based process.

Since we describe a protocol involving a human step rather than a fully automated process, we do not test our protocol on known benchmarks \cite{sessler}. Instead, we test our methodology on: (i) the six problems in the 2025 IMO, (ii) a set of sixty-six open conjectures in number theory taken from the recent paper \cite{cohenprimes}, (iii) a set of conjectures (in three different mathematical fields chosen by us authors) found an LLM agent in a pre-processing step in our protocol.

Comparative success stories in proving theorems using LLMs focus either on solving IMO problems \cite{deeptheorem,site_gemini,openai_xpost}, or conjectures in combinatorial optimization formulated for the explicit purpose of testing the system \cite{feldmanKarbasi}. We note that neither of these are ``open problems'' in the literature. Interestingly, though, the system described in \cite{feldmanKarbasi} gave enough evidence to its authors that one of their own conjectures designed to test their system (which the system could not prove) was more difficult than the authors had envisaged, and is now an open problem.

By contrast, the protocol described in \cite{funsearch} really did advance the knowledge related to the ``cap set problem'', i.e.~finding the largest possible set of vectors in $\mathbb{Z}^n_3$ such that no three of them sum to zero. While the exact answer is only known for a handful of small values of $n$, the methodologies proposed by \cite{funsearch} improved the upper bound for $n=8$.

Overall, we believe that our approach is the first to close open conjectures from a mathematical paper in the literature. Our belief is strengthened by the negative answer provided by ChatGPT (using the \textsf{gpt-5} LLM) to the question, posed on Sept.~23rd, 2025:
\begin{quote}
Excluding IMO problems and ``open problems'' stated by the authors of the paper describing the LLM that solved them, can you find a list of papers or technical reports that report that an LLM actually solved truly open mathematical conjectures, namely conjectures that can be found in a refereed (mathematical) paper?
\end{quote}
The answer (which took 32s of LLM ``thinking'') concluded with the sentence ``under your strict definition the list is empty today.'' Relaxing the criterion to ``nontrivial, peer-reviewed new theorems not previously stated as a conjecture'' the list provided by ChatGPT (after 26s of ``thinking'') reduces to the paper \cite{funsearch} already mentioned. Since the research in this field is very recent and very intense, we have no doubt that there will soon be other systems that can close open conjectures from the literature.

\section{A theorem proving protocol}
\label{s:protocol}
The protocol we propose employs a Test-Time Verify-Revise (TTVR) loop followed by a human verification. TTVR is employed by analogy with the human process of ssolving hard tasks: devise an initial (possibly wrong) solution, check it, correct it, and repeat.

\subsection{Agents}
The TTVR loop in our protocol involves two computational agents that work side-by-side: a \textit{prover} and a \textit{verifier} interact in an alternating fashion. Both agents are \textsf{gpt-5} LLMs that have access to a Python interpreter. The difference between one and the other is determined by the phrasing of system-prompts and user-prompts (more on this in Sect.~\ref{s:prompts}).

Given a theorem statement $T$ (consisting of premises and the conclusion), at the $i$-th iteration of the loop the prover proposes a proof $P_i$, and the verifier either accepts it or rejects it. If the verifier accepts $P_i$ as a valid proof, the loop terminates; if the verifier rejects $P_i$ as invalid, it provides evidence $V_i$ of the issue and the position $p_i$ in the proof $P_i$ where the issue arises. The proof $P_i$, the position $p_i$, and the evidence $V_i$ are passed to the prover as input for the next iteration $i+1$. We set a maximum number $N$ of iterations of this loop, after which termination is enforced even if a proof was not found.

Suppose that the loop terminates at the $n$-th iteration (where $n\le N$) with a proof $P_n$ that is accepted by the verifier. Then the prover is asked to formalize the proof into the formal sentence $\pi$ in a proof assistant programming language (we use \textsf{lean} \cite{lean4} in our implementation). Next, a human checks that all premises of $\pi$ and its conclusion are a correct formal restatement of the premises and conclusion of $P_n$ (which are equal to those of $T$) written in semi-formal natural language. If this human test is successful, $\pi$ is passed to the corresponding proof assistant for the final validity test. If this test is also passed, the proof $P_n$ is deemed to be valid. In all other cases the proof is rejected.

\subsection{Correctness}
\label{s:correctness}
This protocol has an interesting twist, which can be described by means of a remark to the following correctness result. Let $T$ be statement of the theorem being proved, and assume that neither the human nor the proof assistant can ever be wrong in performing their verifications. 
\ificml\begin{proposition}\else\begin{prop}\fi
\label{prop:correctness}
For any pair $(P,\pi)$ where $P$ is a semi-formal natural language proof of $T$ and $\pi$ is a formalized proof of $T$, if $(P,\pi)$ are accepted by the protocol above, then $\pi$ is a valid proof of $T$.
\ificml\end{proposition}\else\end{prop}\fi
\begin{proof}
Since the premises of $P$ and $\pi$ and their conclusions have been certified by the human to have the same formal semantics, and the rest of the proof has been certified by the proof assistant to constitute a valid inference chain from the premises to the conclusion, $\pi$ is a valid formal proof of $T$.
\end{proof}
First, we remark that the encoding of the symbol $T$ is unspecified: it does not matter whether it is expressed in the semi-formal natural language used by human mathematicians and LLMs or in formal language, because the human in the protocol certifies the semantic equivalence of premises and conclusion between $P$ and $\pi$, and the theorem statement $T$ consists exactly of those premises and conclusion. 

Secondly, and most importantly, the almost trivial Prop.~\ref{prop:correctness} was stated to emphasize its only non-trivial feature: \textit{it does not prove that $P$ is a valid proof of $T$, but only that $\pi$ is}. The reason is the hallucinating scenario already considered in Sect.~\ref{s:litrev}. To keep human interaction minimal, we only require that the human should certify conformance of premises and conclusions of $P$ and $\pi$, but not of the whole proof. There is a chance that either $P$, or $\pi$, or both, were actually the fruit of LLM hallucinations. But as long as: (i) the theorem statement $T$ was correctly stated as input to the protocol, (ii) its formalized premises and conclusion in $\pi$ were certified correct by the human, and (iii) the formalized proof was certified correct by the proof assistant, then $\pi$ is a valid proof of $T$, even though the LLM might have hallucinated.

\subsubsection{Hallucinations in the protocol}
\label{s:hallucinations}
We think that the occurrence of a set of LLM hallucinations that impact the prover/verifier loop and the proof assistant code generation is exceedingly rare, but not impossible. We also think that humans, at least those that are conversant with FS and contemporary computer science, tend to trust a proof that is labeled as ``formally certified'' beyond all doubt. It would therefore be catastrophic if a wrong proof of this kind were to be accepted at large. Our protocol is designed to deliver correct formal proofs even in the case of LLM hallucinations. More precisely, most cases of LLM hallucinations will result in failure to deliver any proof; but in the minuscule chance that hallucinations in prover/verifier and \textsf{lean} code generation should deliver a valid formal proof, that proof will be recognized as valid for $T$ by the human and the proof assistant, and therefore returned to the user as valid by our protocol.

A creative way of interpreting this situation is that we recognize that theorem proving is the province of formal languages, but that the creativity necessary to construct (or find?) a proof requires a set of very human skills, many of which manifest themselves in natural language, and remain present in the enormous set of natural language knowledge stored in LLMs. By several accounts (e.g.~Hardy about Ramanujan), mathematicians often derive their ideas for proofs in flashes, illuminations, and even dreams\footnote{By our own experience, the amount of illuminations/dreams leading to good ideas that turn into actual proofs are a tiny minority, but they are precious nonetheless.}. We like to think that if an LLM has an incredibly rare pair of hallucinations leading to a valid formal proof, it might be considered, rather arbitrarily but satisfactorily, as an excellent analogy of the mathematician's illumination. And our protocol saves those proofs. 

\subsubsection{Further remarks}
Here are a few additional remarks about our protocol.
\begin{enumerate}
\item A practical issue of our protocol is the lack of reproducibility. This stems from the fact that the OpenAI pipelines include a probabilistic system for selecting its output in a likely set. In practice, the impact is that one should save all of the valid proofs found by our system, because, in the worst case, they may not be found again. \label{rem:uncertain}
\item The prover/verifier TTVR interaction loop in our protocol is reminiscent of Interactive Proofs (IP) Systems \cite{interactiveproofs}, but for now this is just an analogy rather than a correspondence. Every result in the IP complexity class is based on a probabilistic verifier that can challenge the prover with random bits. Although by remark \ref{rem:uncertain} the verifier is actually probabilistic, by the same token the prover is also probabilistic. We are unsure whether this warrants a complexity study of our protocol under the banner of IP. The fact that we repeat our loop for at most a given number $N$ of times is unlikely to be admissible in IP. 
\item Our protocol could be restricted to proving theorems that can be formally expressed by existential second-order sentences: by Fagin's theorem \cite{fagin}, such ``theorems'' are equivalent to decision problems in the complexity class \textbf{NP}. This would be tantamount to using our protocol to find solutions to instances of \textbf{NP}-complete problems. The complementary view is that of proving propositional tautologies (which is in \textbf{co-NP} \cite{cookReckhow}). We do not think, however, that this specific use of our protocol will be very fruitful, as \textsc{sat} solvers are much better suited to this task. On the other hand, only problems are hard, not single instances: therefore there is some hope that there could exist some instance of an \textbf{NP}-hard problem where traditional methods such as Branch-and-Bound (BB) will take a longer time than our protocol. So far we have not tested this idea.
\item The other well-known link between proofs and algorithms is the Curry-Howard correspondence \cite{curryHoward}: proofs that can be framed in intuitionistic logic (e.g.~no excluded middle arguments) correspond to terminating algorithms. Restricting our protocol to intuitionistic logic might involve convincing the prover agent to exclude the use of certain proof techniques. The benefit of such a restriction might simply be a computable certificate of validity. Currently, we cannot estimate the additional value of such a certificate with respect to the ACCEPT/REJECT bit of output of a proof assistant that can also tackle classical (non-intuitionistic) logic.
\end{enumerate} 

\section{System implementation}
\label{s:impl}
The protocol presented in Sect.~\ref{s:protocol} is similar, albeit not identical, to several other approaches in the literature (see Sect.~\ref{s:litrev}). The only theoretically interesting innovation is the limited interaction with a human at the very end (see Prop.~\ref{prop:correctness} and the remarks below it), but this just ensures a correctness guarantee: it cannot explain the capability of our protocol for solving open conjectures and discovering new theorem statements (see Sect.~\ref{s:results}).

In this section we discuss some implementation features that we believe are key to this success: namely agents and context. The rest of the TTVR implementation is a standard loop coded in Python, and therefore we do not think that it deserves much credit for the system's success. Still, we make some suggestions about different types of agent interactions which might considerably improve the system's efficacy.

\subsection{Agent defined by prompts}
\label{s:prompts}
Our protocol is implemented using the OpenAI API. The LLM we employ is \textsf{gpt-5}. The TTVR loop distinguishes between the first iteration and the subsequent ones by means of different prompts to the various instances of \textsf{gpt-5}. The prompts consist of imperative instructions written in natural language. Each \textsf{gpt-5} instance receives a system prompt and a user prompt through the API (the ChatGPT user interface only allows the specification of the user prompt). Here are more details about the prompts we use.
\begin{itemize}
\item \textbf{Prover} \underline{at the first iteration}: \textit{system prompt}: specifies that this prover should behave like a rigorous mathematician, gives the output format, and prescribes further rules to be used in questions of plane geometry (this is mostly useful for IMO questions), and gives strict guidelines about focusing on the proof itself rather than produce bullet lists of literature review.
\item \textbf{Prover} \underline{at the first iteration}: \textit{user prompt}: passes the theorem statement provided in semi-formal natural language to the prover, repeats some of the recommendations of the system prompt but applied to the proof rather than to general instance behaviour.
\item \textbf{Prover} \underline{at other iterations}: \textit{system prompt}: repeats the recommendations for the prover at the first iteration, and adds the recommendation for: (i) understanding the core idea and structure of the existing proof so far, (ii) fixing the proof based on the verifier feedback with minimal necessary changes if possible, otherwise produce a new correct proof.
\item \textbf{Prover} \underline{at other iterations}: \textit{user prompt}: passes the theorem statement, the previous proof, and the verifier feedback to the prover, and repeats some of the recommendations of the system prompt but applied to the rpoof rather than to general instance behaviour.
\item \textbf{Verifier} \underline{at all iterations}: \textit{system prompt}: specifies that this verifier should behave like a rigorous mathematician, gives the output format, and repeats some of the recommendations already listed for provers.
\item \textbf{Verifier} \underline{at all iterations}: \textit{user prompt}: passes the theorem statement and the current proof to the verifier, and prescribes a YES/NO answer where NO must come with an brief explanations of all logical flaws found, without including suggestions for fixes; flag as wrong all proofs involving appeals to literature status.
\end{itemize}

\subsubsection{Research mode agents}
\label{s:resmode}
So far, we have described a system which, given a theorem statement $T$, attempts to find its proof: this is the default mode of operations. On top of this system we have also built an advanced ``research mode'' that, given the informal description of a mathematical goal, it attempts to find new and interesting theorems towards that goal, and then attempts to prove them.

The research mode is implemented by means of four more agents (i.e.~other \textsf{gpt-5} instances with specific prompts).
\begin{itemize}
\item The \textbf{literature reviewer} is a meticulous research assistant that searches the existing literature for relevant results in the theorem at hand.
\item The \textbf{context preparer} is a senior research assistant that searches for auxiliary references for a difficult open problem.
\item The \textbf{predictor} is a researcher who proposes new conjectures from looking at the literature in a certain field.
\item Given a theorem $T$ and a proof $P$, the \textbf{refiner} simply recognizes whether $P$ proves or disproves $P$; if the latter, the refiner logically inverts $T$ so that $P$ proves $\neg T$. 
\end{itemize}
We note that all of these agents are deployed once only, either as pre-processing or post-processing (\textbf{refiner}) with respect to the TTVR loop. Moreover, the \textbf{context preparer} is also used in the pre-processing of the default mode when deployed on an conjecture rather than a theorem.

\subsection{Advanced agent interactions}
\label{s:interactions}
The simple TTVR protocol described above would certainly benefit from multiple provers and verifiers working in parallel (possibly organised by prover/verifier ``master'' instances). Because of OpenAI subscription limitations and cost, we were only able to allow for two verifiers working sequentially so as to increase the chances of finding faults in the single prover's work. The prompt of the second verifier is a slight variant of that of the first verifier (described in Sect.~\ref{s:prompts} above).

It is not hard to imagine more advanced interaction schemes. Within the same simple TTVR loop, one could deploy more complicated message-passing schemes (through sophisticated prompts) between provers and verifiers. Prompts could even be written automatically by agents conditionally on the progress made on the proof. 

The elementary loop itself could be replaced by a full-fledged algorithm that coordinates the agents' work. As an example, consider the well-known BB algorithm for integer programming based on linear relaxations \cite{land_doig}. Theorems can be relaxed by strenghtening premises or weakening conclusions: by assigning a measure to such relaxations, one could devise a theorem-proving branch-and-bound algorithm that automatically devises a decomposition of the theorem, thereby making it easier to derive a proof. This would yield a form of nontrivial recursion proof for the given theorem.

Unfortunately (for us), implementing, fine-tuning, and testing such schemes would require an exceedingly costly subscription to OpenAI. We therefore encourage OpenAI researchers to consider the ideas in this section\footnote{Or, alternatively, to either employ us for this purpose or at least give us free unlimited subscriptions!}.

\subsection{LLM Context}
According to the OpenAI API documentation, each \textsf{gpt-5} instance has a context window of four-hundred thousand tokens (the word ``token'' is to be interpreted in the light of natural language parsing). If we used a single \textsf{gpt-5} instance for all provers and verifiers at every iteration we would quickly fill the context window, which might lead to degraded performance and increase the chances of hallucinations \cite{michelangelo}: therefore each of our agents is used once and then discarded; new agents are created in the subsequent TTVR loop iterations.

Moreover, since the system has to correct its own wrong proofs, we think that the context bias might prevent the prover from exploring new ideas. For these reasons, every instance of \textsf{gpt-5} is discarded after use. Every prover and verifier at each iteration is a new instance.

\subsection{Human interaction issues}
\label{s:leanIssues}
In practice, the human interaction, albeit limited to a verification at the end, remains very time-consuming. This is mostly due to the limited ability shown by \textsf{gpt-5} in converting its own natural language proofs into formal \textsf{lean} codes. So, although our protocol does minimize human interaction, this interaction remains considerable, because the human has to construct the whole \textsf{lean} code from a relatively thin base. A more intensive LLM training using \textsf{lean} codes could improve the situation. This is another situation (as that described in Sect.~\ref{s:interactions}) which would be better tackled by OpenAI.

In order to reduce human interaction, we had initially instructed OpenAI's Codex interface to reduce to ``axiom'' any proof step that it could find by means of a web search. This reduced the extent of \textsf{lean} checking (the formalization was partial) but made it possible to run a limited certification step at the end of our protocol\footnote{We have since enlisted the help of human experts in analytic number theory to help us verify the solutions of conjectures in \cite{cohenprimes}. Some of them were found to be incorrect despite the limited formalization effort; and not all of them have been yet vetted.}.

\section{Results}
\label{s:results}
In this section we give an account of our system's achievements. We note that \textsf{gpt-5} was trained on data up to 2024, while the problems we submitted to our system all date from 2025. Therefore, the LLM in our TTVR loop cannot have been contaminated in the sense of \cite{site_matharena_ai}, namely that its training set cannot contain any IMO 2025 information, nor any solution to conjectures in \cite{cohenprimes} (the situation for the research mode conjectures in Sect.~\ref{s:results:resmode} is more fluid in this respect: some contamination could be possible). We also emphasize the fact that our prompts do not include instructions to the LLM to the effect of checking the literature; on the contrary, they discourage the production of bullet lists with literature references. 

Given that our system is based on a TTVR loop, we consider the number of iterations required to arrive at a solution as an index of difficulty of the problem for our LLM-based protocol. Accordingly, we compiled some tables in this sense (Tables \ref{t:imoitn}-\ref{t:sphpack_itn}).

The number of iterations $N$ for the TTVR loop was fixed at $15$ for the whole set of experiments. Another constant $M=5$ was used to abort the loop after an excessive number of ``bad gateway'' error messages from the OpenAI API: typically, harder problems run into longer reasoning time, which increases the chances of such API errors. 

\subsection{2025 IMO Problems}
\label{s:imo25}
The 2025 edition of the IMO contained six problems. Our system solved five out of six problems (it could not solve the sixth one), which is the same achievement announced by the specialized and undisclosed LLMs used by OpenAI \cite{openai_xpost} and Google DeepMind \cite{site_gemini} for the task. Both problem statements and solutions can be obtained from the PDF document at \href{https://web.evanchen.cc/exams/IMO-2025-notes.pdf}{\texttt{evanchen.cc}}. The length of the human proofs of these problems have lengths reported in the second row (labelled ``human'') of Table \ref{t:imoprooflen}.
\begin{table}[!ht]
\begin{center}
\begin{tabular}{l||r|r|r|r|r|r}
problem & 1 & 2 & 3 & 4 & 5 & 6 \\ \hline
human & 321 & 495 & 320 & 599 & 434 & 639 \\
LLM & 1105 & 490 & 604 & 690 & 434 & NA
\end{tabular}
\end{center}
\caption{Human versus generated proof lengths for the IMO 2025 problems in number of tokens. Drawings in human proofs were not counted in their lengths.}
\label{t:imoprooflen}
\end{table}
The human proof lengths are expressed in number of space-separated tokens extracted by the ASCII representation of the \href{https://web.evanchen.cc/exams/IMO-2025-notes.pdf}{PDF} referred to above. The LLM-generated proof lengths are expressed in number of space-separated tokens in the proofs given in MarkDown language \cite{markdown} with inline LaTeX.

\subsubsection{Notes about the proofs}
We now look at the five proofs of problems 1--5 in more detail.
\begin{enumerate}[Problem 1.]
\item The human and computer-generated proofs are different, and give rise to very different proof lengths in Table \ref{t:imoprooflen}. Our system failed to consider a shortcut taken by the human proof.
\item The two proofs are different, even though the proof lengths are similar. The human proof is more compact and geometrically clearer to a human reader. The computer proof is algebraic (instead of geometric) insofar as it uses the complex plane to reason about plane geometry (this was due to an explicit instruction to the LLM on our part through a prompt). It relates to the human proof similarly to the way Miles Edwards' proof of Heron's formula relates to most other proofs of the same result (see \cite[\S 2]{six} for more details).
\item Overall, the structure of the two proofs is the same; the human proof is more compact but also leaves more to be worked out by the reader w.r.t.~the computer proof, which explains the proof length difference. The case analysis at the end differs in the details.
\item The two proofs are almost the same. Both proofs state the solution nonconstructively, and then prove its correctness. The differ only for some minor details (consistently with the similar proof lenghts).
\item The two proofs are essentially the same, but differ on a few details (the identical lengths are just a coincidence). Both proofs state the solution nonconstructively, and then prove its correctness. The human proof looks ``neater'' to a human reader, but the computer proof gives more details, which makes the inferences easier to understand. There is a point at which the human proof infers correctnes from the fact that the value of a certain variable belongs to a given closed real interval having zero lower bound; in the same situation, the computer proof simply assigns the zero value to the variable. In this case it seems that the human proof contributes more knowledge even if that knowledge is not useful to the proof itself, while the computer proof states the weakest premise for the inference. Lastly, the fact that there is a unique value of the main game parameter for which there is no winning strategy for either player is stated by both proofs, but it is only explicitly proved by the computer proof. 
\end{enumerate}

\subsubsection{Qualitative comparison of our system compared to others}
As mentioned above, we use the number of TTVR iterations as an index of problem difficulty for our system.
\begin{table}[!ht]
\begin{center}
\begin{tabular}{l||r|r|r|r|r|r}
problem   & 1 & 2 & 3 & 4 & 5 & 6 \\ \hline
$|$itn$|$ & 9 & 7 & 3 & 6 & 5 & NA \\
correct?  & Y & Y & Y & Y & Y & NA \\
certified?&   &   & Y &   &   & NA
\end{tabular}
\end{center}
\caption{Number of iterations required by the TTVR loop to arrive at a solution of the IMO25 problems, together with a human evaluation of correctness and a \textsf{lean} certification.}
\label{t:imoitn}
\end{table}
According to Table\footnote{Missing entries correspond to ongoing tests.} \ref{t:imoitn}, the easiest problem for humans (Problem 1) is the hardest for our protocol; the second-hardest is Problem 2. We note that both problems are geometric. The benchmark found in \cite{site_matharena_ai} agrees on Problem 6 (no LLM-based approach could solve it), and approximately also on the relative ease with which such approaches can solve Problems 3, 4, and 5. This benchmark tells a rather different story for Problems 1 and 2: Problem 2 is much more difficult than Problem 1 --- in fact it is the most difficult problem in the set 1--5. Our protocol, however, finds that the opposite holds: Problem 1 is harder than problem 2 (admittedly according to a different measure than that of the benchmark). In general, given that LLMs are based on symbolic textual training more than on geometric drawings, one might guess that LLMs find geometric problems harder to solve than other types of problems. Our tests provide one more piece of evidence to reinforce confidence in this guess. 

\subsection{Conjectures of primes and cyclics}
\label{s:cyclics}
We considered number theory as a promising field to test our system for several reasons: (i) it is a cornerstone of pure mathematics; (ii) many of its theorems and conjectures, albeit difficult, can be formalized using first-order or second-order logic; (iii) many of its theorem statements and conjectures can be understood by people possessing limited mathematical education. It promised to be a hard benchmark with potentially wide audience.

For this particular test we wanted to benchmark the ability of our system to prove stated theorems rather than its creativity to discover new theorems. But instead of conceiving our own conjectures (as \cite{feldmanKarbasi}) we thought the exercise would be less biased if we used existing ones. While perusing the OEIS \cite{oeis} we came across a conjecture which led us to \cite{cohenprimes}, an article that lists 66 conjectures about primes and cyclic numbers. 

Cyclic numbers are analogous to prime numbers in a certain sense, and therefore it makes sense to conjecture that many results known for primes should also apply to cyclics. The definition of a \textit{cyclic number} is as follows. Let $n$ be a positive integer, and consider the cardinality $m$ of the set $C$ of positive integers $r<n$ and relatively prime to $n$ (obviously, $m$ is a function of $n$). If $n,m$ are relatively prime, then $n$ is cyclic. It is analogous to the definition of ``prime'' insofar as they both involve the comparison of an integer with the cardinality $m$ of a set of numbers $r<n$ that have a certain relationship with $n$ concerning integer division. If the set is that of the divisors of $n$ and $m=1$ (only $r=1<n$ divides $n$), then $n$ is prime; if the set is that of the coprimes of $n$ and $\mathrm{gcd}(n,m)=1$, then $n$ is cyclic. To render notation more compact, we introduce Euler's \textit{totient} function $\phi(n)=m$ to be the cardinality of the integers $r<n$ and coprimes with $n$, and define $n$ to be cyclic if $\mathrm{gcd}(n,\phi(n))=1$. Note that, if $n$ is prime, then $r$ is coprime with $n$ for each $r<n$, so $m=n-1$, and since $\mathrm{gcd}(n,n-1)=1$ for each integer $n$, each prime is also cyclic. The paper \cite{cohenprimes} goes on to list the 66 (numbered) conjectures about cyclic numbers.

Our system produced a proof\footnote{In analytic number theory, a ``proof'' can only be correct, otherwise it is not a proof, but a fallacious argument. An ``argument'' may be correct or incorrect, and a correct argument provides a proof. Moreover, one does not use ``theorem'' to just indicate the theorem statement: a theorem is the statement followed by its proof. Because the subject matter of this paper is the automated generation of proofs rather than analytic number theory, we abuse those rules to keep descriptions more compact: proofs and refutations may be correct or incorrect.} or refutation for twenty out of the sixty-six conjectures: namely 3, 6, 9, 14, 17, 20, 32, 35, 36, 37, 41, 42, 47, 52, 53, 54, 56, 59, 60, 61, with seventeen proofs and three refutations (35, 53, 59). Seven out of the sixty-six conjectures, namely 2, 3, 4, 36, 37, 52, 53, had already been closed by C.~Pomerance \cite{pomerance} prior to the publication of \cite{cohenprimes}, by personal communication to its author. The five conjectures solved by Pomerance that appear in our list (3, 36, 37, 52, 53) must therefore be removed from the list of ``open conjectures''. This leaves our system with 17 (previously open) conjectures for which it obtained a proof.
\begin{table}[!ht]
\begin{center}
\begin{tabular}{l||@{\hspace{3pt}}r@{\hspace{3pt}}|@{\hspace{3pt}}r@{\hspace{3pt}}|@{\hspace{3pt}}r@{\hspace{3pt}}|@{\hspace{3pt}}r@{\hspace{3pt}}|@{\hspace{3pt}}r@{\hspace{3pt}}|@{\hspace{3pt}}r@{\hspace{3pt}}|@{\hspace{3pt}}r@{\hspace{3pt}}|@{\hspace{3pt}}r@{\hspace{3pt}}|@{\hspace{3pt}}r@{\hspace{3pt}}|@{\hspace{3pt}}r@{\hspace{3pt}}|@{\hspace{3pt}}r@{\hspace{3pt}}|@{\hspace{3pt}}r@{\hspace{3pt}}|@{\hspace{3pt}}r@{\hspace{3pt}}|@{\hspace{3pt}}r@{\hspace{3pt}}|@{\hspace{3pt}}r@{\hspace{3pt}}|@{\hspace{3pt}}r@{\hspace{3pt}}|@{\hspace{3pt}}r@{\hspace{3pt}}|@{\hspace{3pt}}r@{\hspace{3pt}}|@{\hspace{3pt}}r@{\hspace{3pt}}|@{\hspace{3pt}}r}
conjecture & 3 & 6 & 9 & 14 & 17 & 20 & 32 & 35 & 36 & 37 & 41 & 42 & 47 & 52 & 53 & 54 & 56 & 59 & 60 & 61 \\ \hline
$|$itn$|$ & 4 & 9 & 9 & 6 & 1 & 9 & 3 & 3 & 4 & 8 & 15 & 1 & 9 & 9 & 1 & 9 & 9 & 1 & 1 & 1 \\
\textit{O} or \textit{C} & C & O & O & O & O & O & O & O & C & C & O & O & O & C & C & O & O & O & O & O \\
\textit{P} or \textit{R} & P & P & P & P & P & P & P & R & P & P & P & P & P & P & R & P & P & R & P & P \\
correct? & ? & Y & Y & N & Y & Y & Y & N & ? & ? & Y & Y & Y & ? & ? & Y & Y & Y & Y & Y \\
certified? & ? & Y & Y & Y & Y & & & & ? & ? & Y & & & ? & ? & & & & & 
\end{tabular}
\end{center}
\caption{Number of iterations required by our TTVR loop to produce either correct and \textsf{lean}-certified or incorrect \textit{P}roofs or \textit{R}efutations of the conjectures in \cite{cohenprimes}.}
\label{t:cyclics_itn}
\end{table}
In Table \ref{t:cyclics_itn} we report, for each proof produced\footnote{Missing entries in Table \ref{t:cyclics_itn} correspond to ongoing tests.} by our system: the number of TTVR iterations ($|\mbox{itn}|$), whether the conjecture was open or already closed (\textit{O}/\textit{C}), whether the conjecture was settled by proof or refutation (\textit{P}/\textit{R}), whether the authors of this paper (who are not specialists in analytic number theory\footnote{These proofs ar enow being vetted by experts in the subject matters, who have already invalidated some of the proofs we believed to be correct, and have yet to go through the complete list.}) think that the semi-formal natural language proofs are correct or not, and whether the \textsf{lean} certificate is correct and conformant with the premises and conclusions of the corresponding proofs.

We note that some of the confirmed conjectures have been ``closed'' only partially, i.e.~in the asymptotic sense, not in the absolute one. For example, the proof of Conjecture 6 shows that there is an $n_0$ such that for every $n>n_0$ the conjecture holds, but we have no idea about this $n_0$, so Conjecture 6 is still open ``for all $n\in\mathbb{N}$''. This is also the case for Conj.~32.

In one of the refuted conjectures, namely Conj.~59, our system has provided a formally correct refutation that would be unsatisfactory to any human mathematician: it addresses one of finitely many exceptions, providing no refutation for the asymptotic part.

The most difficult conjecture to solve, according to the number of TTVR iterations\footnote{The ``?'' entries in Table \ref{t:cyclics_itn} are due to a mishap in our file management: we are re-running these tests.}, is 41. We report it here as an example, as Thm.~\ref{thm:cohen41}: $c_n$ is the $n$-th cyclic number, and $\log$ is the natural logarithm.

\begin{greyblock}
\ificml\begin{theorem}[Third Firoozbakht conjecture on cyclics]\else\begin{thm}[Third Firoozbakht conjecture on cyclics]\fi
\label{thm:cohen41}
For every $k\in \mathbb{N}$, there exists a least $m\in\mathbb{N}$, call it $N(k)$, such that, for all $n>N(k)$, we have $c_n^{1/(n+k)}>c_{n+1}^{1/(n+k+1)}$.
\ificml\end{theorem}\else\end{thm}\fi
\begin{proof}
Let $L(x)=\log_3 x=\log\log\log x$ for $x\ge e^{e^e}$ and let
\[C(x)=\#\{m\le x: \gcd(m,\varphi(m))=1\}.\]
By Pollack's refinement of Erd\H{o}s' asymptotic, there exist absolute constants $X_0\ge e^{e^e}$ and $A_0>0$ such that for all $x\ge X_0$,
\[
C(x)=e^{-\gamma}x\Bigl(\frac1{L(x)}-\frac{\gamma}{L(x)^2}+\frac{q}{L(x)^3}+R(x)\Bigr),\qquad q=\gamma^2+\frac{\pi^2}{12},\quad |R(x)|\le \frac{A_0}{L(x)^4}.
\]
Define the smooth comparison functions
\[
F_\pm(x)=e^{-\gamma}x\Bigl(\frac1{L(x)}-\frac{\gamma}{L(x)^2}+\frac{q}{L(x)^3}\pm\frac{A_0}{L(x)^4}\Bigr)\qquad(x\ge X_0),
\]
so that for $x\ge X_0$,
\[
F_-(x)\le C(x)\le F_+(x).
\]

1) \textit{Uniform lower bound for $F_-'$}. Write $\ell=L(x)$ and $G(\ell)=\ell^{-1}-\gamma\ell^{-2}+q\ell^{-3}-A_0\ell^{-4}$. Since $L'(x)=(x\log x\,\log_2 x)^{-1}$,
\[
F_-'(x)=e^{-\gamma}\Bigl[G(\ell)+x\,G'(\ell)L'(x)\Bigr].
\]
Because $G'(\ell)=-\ell^{-2}+O(\ell^{-3})$, there exist $X_1\ge X_0$ and $C_1>0$ such that for all $x\ge X_1$,
\[
\bigl|x\,G'(\ell)L'(x)\bigr|\le \frac{C_1}{\ell^2\log x\,\log_2 x}.
\]
Moreover $G(\ell)=\ell^{-1}+O(\ell^{-2})$. As $\ell\to\infty$ and $\log x\,\log_2 x\to\infty$, enlarging $X_1$ if needed we obtain
\[
F_-'(x)\ge e^{-\gamma}\Bigl(\frac{1}{2L(x)}\Bigr)\qquad(x\ge X_1).\tag{1}
\]
In particular $F_-$ is strictly increasing on $[X_1,\infty)$ and, since $F_-(x)\gg x/L(x)$, one has $F_-(x)\to\infty$ as $x\to\infty$.

2) \textit{One-step growth via level-crossing of $F_-$}. Fix $n$ with $c_n\ge X_1$ and set $y=c_n$, so $C(y)=n$ and $F_-(y)\le n$. Because $F_-$ is continuous, strictly increasing, and unbounded, there is a unique $\Delta(y)\ge0$ such that
\[
F_-(y+\Delta(y))=n+1.
\]
Then $C(y+\Delta(y))\ge F_-(y+\Delta(y))=n+1$, hence the first point where $C$ reaches $n+1$ (namely $c_{n+1}$) lies in $(y,\,y+\Delta(y)]$. Thus
\[
0<c_{n+1}-c_n\le \Delta(y).
\]
By the mean value theorem there exists $\xi\in[y,\,y+\Delta(y)]$ with
\[
F_-(y+\Delta(y)) - F_-(y)=F_-'(\xi)\,\Delta(y).
\]
Because $F_-(y+\Delta(y))=n+1\ge n\ge F_-(y)$, the left-hand side is $\ge1$, so by (1)
\[
\Delta(y)\le \frac{1}{F_-'(\xi)}\le 2e^{\gamma}L(\xi).\tag{2}
\]
As $L(t)=o(t^{\varepsilon})$ for any fixed $\varepsilon>0$, there exists $X_2\ge X_1$ such that $2e^{\gamma}L(t)\le t/4$ for all $t\ge X_2$. Suppose $y\ge X_2$. If $\Delta(y)\ge y$, then from (2) we get $\Delta(y)\le(y+\Delta(y))/4$, i.e. $3\Delta(y)\le y$, a contradiction. Hence $\Delta(y)<y$, so $y+\Delta(y)\le2y$ and, by monotonicity of $L$ and the elementary bound for $t\ge e^{e^e}$,
\[
L(y+\Delta(y))\le L(2y)\le L(y)+\log 2\le 2L(y).
\]
Combining with (2) yields
\[
0<c_{n+1}-c_n\le\Delta(y)\le 4e^{\gamma}L(y)=4e^{\gamma}L(c_n)\qquad(n\text{ large}).\tag{3}
\]

3) \textit{A coarse upper bound for $L(c_n)/c_n$}. From $C\le F_+$ and, for large $\ell=L(x)$, the estimate $\ell^{-1}-\gamma\ell^{-2}+q\ell^{-3}+A_0\ell^{-4}\le 2\ell^{-1}$, we obtain
\[
C(x)\le \frac{2e^{-\gamma}x}{L(x)}\qquad(x\ge X_3)
\]
for some $X_3\ge X_2$. Evaluating at $x=c_n\ge X_3$ gives
\[
\frac{L(c_n)}{c_n}\le \frac{2e^{-\gamma}}{n}. \tag{4}
\]

4) \textit{Bounding the logarithmic increment.} From Eq.~(3)-(4), using $\log(1+u)\le u$,
\[
\log\frac{c_{n+1}}{c_n}\le \frac{c_{n+1}-c_n}{c_n}\le 4e^{\gamma}\frac{L(c_n)}{c_n}\le \frac{8}{n}\qquad(n\text{ large}).\tag{5}
\]

5) \textit{Conclusion}. Let $a_n=\log c_n$. Since $c_n\ge n$, we have $a_n\ge\log n$. Fix $k\in\mathbb N$. Choose $N(k)$ so large that for all $n\ge N(k)$: (i) $n\ge2k$, (ii) $c_n\ge X_3$, and (iii) $\log n>16$. Then by (5), for all such $n$,
\[
(n+k)(a_{n+1}-a_n)\le \frac{n+k}{n}\cdot 8\le 16<\log n\le a_n.
\]
This is equivalent to
\[
\frac{\log c_n}{n+k}>\frac{\log c_{n+1}}{n+k+1},
\]
i.e. $c_n^{1/(n+k)}>c_{n+1}^{1/(n+k+1)}$. As $k$ was arbitrary, the claim follows.
\end{proof}
\end{greyblock}

\subsection{Research mode results}
\label{s:results:resmode}
We used our system in research mode (Sect.~\ref{s:resmode}) on three different goals (literally specified to the system by the sentences that follow):
\begin{enumerate}
\item improve gradient descent in machine learning;
\item the 1-2-3 conjecture in graph theory;
\item sphere packing in Euclidean spaces.
\end{enumerate}
The first (pre-processing) step is to ask the LLM to construct a list of ``seed results'' from the single sentence given by the human user (e.g.~one of the three above sentences). Second, we deploy the literature reviewer agent that returns interesting conjectures, problems, theorem statements, or other provable sentence from sources in the open literature. Third, the context preparer returns a relevant selection of formal statements to be proved, by discarding some of the items from the literature reviewer (specifically, items for which the solution is already known). The TTVR loop is then deployed on each of them in turn.

In the rest of this section, output generated from the system (possibly only slightly edited for \LaTeX\ typos or to merge multiple definitions) is on light gray background\footnote{None of the proofs in this section has yet been verified formally in \textsf{lean}, nor has it been verified by human experts in the subject matters of the three topics.}.

\subsubsection{Gradient descent in ML}
\label{s:gdml}
The human-provided guideline ``improve gradient descent in machine learning'' used as input for the pre-processing step yielded the following seed results.
\begin{greyblock}
\begin{enumerate}
\item Let $f$ be convex and $L$-smooth. Gradient descent with step size $\eta \in (0, 1.75/L]$ induces a convex optimization curve: the mapping $n \mapsto f(x_n)$ is convex (equivalently, the sequence $f(x_n) - f(x_{n+1})$ is non-increasing).
\item For any $L > 0$ there exists a convex $L$-smooth function and initialization such that for every step-size $\eta \in (1.75/L, 2/L)$, the gradient descent optimization curve is not convex (despite converging and monotonically decreasing).
\item  For convex $L$-smooth $f$ and $\eta \in (0, 2/L]$, the sequence of gradient norms $\{\|\nabla f(x_n)\|\}$ is non-increasing.
\item For convex $L$-smooth $f$, the gradient flow optimization curve $t \mapsto f(x(t))$ is always convex.
\item For convex $L$-smooth $f$, under gradient flow, the function $t \mapsto \|\nabla f(x(t))\|$ is non-increasing.
\item Fix any $L>0$ and any relative noise level $\delta\in(0,1)$. There does not exist a positive universal step size $\eta_{\max}(\delta,L)>0$ with the property that for every (even one-dimensional) convex $L$-smooth quadratic $f$, every nonzero initialization $x_0\ne 0$, and every sequence of inexact-gradient noises satisfying $\|e_n\|\le\delta\|\nabla f(x_n)\|$, the inexact gradient descent iterates $x_{n+1}=x_n-\eta(\nabla f(x_n)+e_n)$ produce a convex sequence of function values $n \mapsto f(x_n)$ for every choice of step size $0<\eta\le\eta_{\max}(\delta,L)$. In particular, no nonzero universal convexity-preserving step-size depending only on $\delta$ and $L$ (and valid for all one-dimensional convex $L$-smooth quadratics and all nonzero initializations) exists when $\delta>0$.
\end{enumerate}
\end{greyblock}

The literature reviewer found 19 interesting sources from the literature\footnote{Many of these sources are arXiv reports, so they may never have been peer-reviewed.} leading to 19 conjectures. The context preparer agent formulated a set of 14 conjectures of interest that we report below.
\begin{greyblock}
\begin{enumerate}[Conjecture 1.]
\item (EG convexity up to the stability limit.) Let $f$ be convex and $L$-smooth. For extragradient with any stepsize $\eta\in(0,2/L]$, the optimization curve is convex: the sequence $n\mapsto f(x_n)$ satisfies $\delta_n\ge 0$ for all $n$.
\item (Two-point time-averaging restores convexity beyond 1.75.) Let $f$ be convex and $L$-smooth and fix $\eta\in(0,2/L]$. There exists a sharp threshold $\lambda_\star(\theta)\in[0,1/2)$ with $\lambda_\star(\theta)=0$ for $\theta\le 1.75$ and $\lambda_\star(\theta)\downarrow0$ as $\theta\downarrow1.75$ such that for all $\lambda\in[\lambda_\star(\theta),1-\lambda_\star(\theta)]$ the 2-term averaged curve $n\mapsto M^{(\lambda)}(a)_n=(1-\lambda)f(x_n)+\lambda f(x_{n+1})$ is convex for every convex $L$-smooth $f$ and every initialization, while for any fixed $\lambda\notin[\lambda_\star(\theta),1-\lambda_\star(\theta)]$ there exist counterexamples.
\item (Implicit step convexifies the discrete curve.) Let $f$ be convex and $L$-smooth. For the implicit (backward Euler) method $x_{n+1}=x_n-\eta\nabla f(x_{n+1})$ with any $\eta>0$, the sequence $n\mapsto f(x_n)$ is convex: $\delta_n\ge 0$ for all $n$. Moreover, if $f$ is $\mu$-strongly convex, then $(\Delta_n)$ is strictly decreasing unless $x_n$ is optimal.
\item (Gradient-norm convexity in the conservative regime.) Let $f$ be convex and $L$-smooth. For GD with $\eta\in(0,1/L]$, the sequence $n\mapsto\|\nabla f(x_n)\|^2$ is convex: $\|\nabla f(x_{n+2})\|^2-2\|\nabla f(x_{n+1})\|^2+\|\nabla f(x_n)\|^2\ge 0$ for all $n$.
\item (Convexity of a Moreau-smoothed evaluation along GD.) Let $f$ be convex and $L$-smooth and fix $\eta\in(0,2/L]$. There exists a universal $\bar\gamma(\theta)\in(0,\infty)$ with $\bar\gamma(\theta)=\Theta\!\big((2-\theta)/L\big)$ as $\theta\uparrow2$ such that, for every $\gamma\in(0,\bar\gamma(\theta)]$, the sequence $n\mapsto e_\gamma(f)(x_n^{\text{GD}})$ is convex in $n$.
\item (Windowed convexity for GD near the stability edge.) Let $f$ be convex and $L$-smooth and fix $\eta\in(0,2/L]$. Then there exists an integer window size $W_\star(\theta)=O\big(1/(2-\theta)\big)$ such that the $W_\star$-point running average $b_n=\tfrac{1}{W_\star}\sum_{j=0}^{W_\star-1} f(x_{n+j})$ is convex in $n$.
\item (Adaptive curvature-aware steps ensure convexity up to $2/L$.) Let $f$ be convex and $L$-smooth, and suppose along the GD trajectory the local smoothness surrogate $L_n=\sup_{t\in[0,1]}\|\nabla^2 f(x_n+t(x_{n+1}-x_n))\|$ satisfies $L_{n+1}\le L_n$ for all $n$ (nonincreasing local curvature). Then for any constant $\eta\in(0,2/L_0]$, the GD loss curve is convex: $\delta_n\ge 0$ for all $n$.
\item (Exact line-search GD yields convex values.) Let $f$ be convex and $L$-smooth. For exact line-search along negative gradient, $x_{n+1}=x_n-\eta_n\nabla f(x_n)$ with $\eta_n=\operatorname*{argmin}_{\eta\ge0} f(x_n-\eta\nabla f(x_n))$, the sequence $n\mapsto f(x_n)$ is convex for all initializations.
\item (Moreau-envelope descent curve convexity under GD on $f$.) Let $f$ be convex and $L$-smooth, and fix $\eta\in(0,2/L]$. Let $y_n=x_n-\eta\nabla f(x_n)$ be the GD predictor and define $m_n=e_{\eta}(f)(y_n)$. Then $n\mapsto m_n$ is convex for all $n$.
\item (EG monotonicity of one-step decrease.) Let $f$ be convex and $L$-smooth. For extragradient with any $\eta\in(0,2/L]$, the one-step decreases are nonincreasing: $\Delta_{n+1}\le \Delta_n$ for all $n$.
\item (Diagonal preconditioning criterion.) Let $f$ be convex and coordinate-wise $L_i$-smooth (i.e., $\partial_i\nabla f$ is $L_i$-Lipschitz). Consider preconditioned GD $x_{n+1}=x_n-D\nabla f(x_n)$ with diagonal $D=\operatorname{diag}(d_i)$, $d_i>0$, and define $\Theta=\max_i L_i d_i$. If $\Theta\in(0,1.75]$, then for every convex coordinate-wise smooth $f$ and any initialization, the optimization curve $n\mapsto f(x_n)$ is convex; furthermore, for every $\Theta\in(1.75,2)$, there exist convex coordinate-wise smooth $f$ and initializations where convexity fails even though $f(x_n)\searrow f_*$.
\item (Lookahead with small pull guarantees convexity.) Let $f$ be convex and $L$-smooth. Consider lookahead with inner GD step-size $\eta\in(0,2/L]$ and outer parameters $k\in\mathbb{N}$, $\alpha\in(0,\bar\alpha(\theta,k)]$ with $z_{t+1}=(1-\alpha)z_t+\alpha x_{t,k}$, where $x_{t,0}=z_t$ and $x_{t,j+1}=x_{t,j}-\eta\nabla f(x_{t,j})$. There exists $\bar\alpha(\theta,k)>0$ with $\bar\alpha(\theta,k)=\Theta\!\big((2-\theta)/k\big)$ as $\theta\uparrow2$ such that $t\mapsto f(z_t)$ is convex for every convex $L$-smooth $f$ and initialization.
\item (Implicit step is a discrete convexifier for GD.) Fix convex $L$-smooth $f$ and any $\eta\in(0,2/L]$. Define the hybrid scheme $x_{n+1}=x_n-\eta\nabla f(x_n)$, $z_{n+1}=z_n-\eta\nabla f(z_{n+1})$ with $z_0=x_0$, and the mixed curve $c_n=\tfrac12(f(x_n)+f(z_n))$. Then $n\mapsto c_n$ is convex for all $n$.
\item (Restart-on-curvature rule ensures convexity without $L$.) Let $f$ be convex with $L$-Lipschitz gradient unknown. Consider GD with an adaptive stepsize $\eta_n$ obtained by doubling until $\Delta_{n-1}\ge\Delta_n$, and halving otherwise. This rule produces a piecewise-constant stepsize sequence with finitely many halving events and guarantees global convexity of $n\mapsto f(x_n)$.
\end{enumerate}
\end{greyblock}

The TTVR loops deployed on each managed to settle 5 out of the 14 statements, and, more precisely, statements 3, 4, 5, 7, and 13. Of these, 3 and 13 were proved while 4, 5, and 7 were refuted. We record the number of TTVR iterations for each of the successful runs in Table \ref{t:cvxopt_itn}.
\begin{table}[!ht]
\begin{center}
\begin{tabular}{l||r|r|r|r|r}
conjecture & 3 & 4 & 5 & 7 & 13 \\ \hline
$|$itn$|$ & 2 & 1 & 1 & 5 & 6
\end{tabular}
\end{center}
\caption{Number of iterations required by our TTVR loop to arrive at a solution of the convex optimization conjectures.}
\label{t:cvxopt_itn}
\end{table}
The table shows that the system found conjecture 13 to be the hardest. We report it here by way of example (exactly as the system produced it, with a couple of minor cosmetic changes). We only need to introduce the slightly non-standard definition of ``convex sequence'': a sequence $c_n$ is \textit{convex} if the sequence of first differences is non-decreasing, i.e.~$(c_{n+2}-c_{n+1})\ge(c_{n+1}-c_n)$ for all indices $n$ over which the sequence is defined. 

\begin{greyblock}
\ificml\begin{theorem}\else\begin{thm}\fi
\label{thm:cvxopt13}
Let $f: \mathbb{R}^d\to\mathbb{R}$ be convex and differentiable, and fix any stepsize $\eta>0$. Define the explicit and implicit iterates by \(x_{n+1}=x_n-\eta\nabla f(x_n)\) and \(z_{n+1}=z_n-\eta\nabla f(z_{n+1})\) with arbitrary initial points \(x_0,z_0\), and set \(c_n=\tfrac12\big(f(x_n)+f(z_n)\big)\). Then the sequence \(n\mapsto c_n\) is convex, i.e., \(c_{n+2}-2c_{n+1}+c_n\ge 0\) for all \(n\).
\ificml\end{theorem}\else\end{thm}\fi
\begin{proof}
Let $f:\mathbb{R}^d\to\mathbb{R}$ be convex and differentiable, and fix $\eta>0$. Define the explicit and implicit iterates by
\[x_{n+1}=x_n-\eta\nabla f(x_n),\qquad z_{n+1}=z_n-\eta\nabla f(z_{n+1}),\]
with arbitrary initial points $x_0,z_0$, and set $c_n=\tfrac12\big(f(x_n)+f(z_n)\big)$. We prove $\delta^c_n=c_{n+2}-2c_{n+1}+c_n\ge 0$ for all $n$.

Write the Bregman divergence $D_f(u,v)=f(u)-f(v)-\langle\nabla f(v),u-v\rangle\ge 0$. Denote $g_n=\nabla f(x_n)$, $h_{n+1}=\nabla f(z_{n+1})$, and the steps $\Delta x_n=x_{n+1}-x_n=-\eta g_n$, $\Delta z_{n+1}=z_{n+1}-z_n=-\eta h_{n+1}$. The identities (obtained from the definition of $D_f$) are
\[\Delta^x_n=f(x_n)-f(x_{n+1})=\frac{1}{\eta}\|\Delta x_n\|^2- D_f(x_{n+1},x_n)\]
and
\[\Delta^z_n=f(z_n)-f(z_{n+1})=\frac{1}{\eta}\|\Delta z_{n+1}\|^2+D_f(z_n,z_{n+1}).\]
Since $\Delta^c_n=c_n-c_{n+1}=\tfrac12(\Delta^x_n+\Delta^z_n)$, we have
\[2\,\delta^c_n=(\Delta^x_n-\Delta^x_{n+1})+(\Delta^z_n-\Delta^z_{n+1})=\eta\,S_g+S_B,\]
where
\begin{eqnarray*}
S_g &=& \big(\|g_n\|^2-\|g_{n+1}\|^2\big)+\big(\|h_{n+1}\|^2-\|h_{n+2}\|^2\big) \\
S_B &=& \big(D_f(x_{n+2},x_{n+1})-D_f(x_{n+1},x_n)\big)+\big(D_f(z_n,z_{n+1})-D_f(z_{n+1},z_{n+2})\big).
\end{eqnarray*}
Use the three-point identity (valid for all $a,b,c$):
\[D_f(a,b)-D_f(b,c) = D_f(a,c)-\langle\nabla f(b)-\nabla f(c),\,a-b\rangle.\]
With $(a,b,c)=(x_{n+2},x_{n+1},x_n)$ and $(a,b,c)=(z_n,z_{n+1},z_{n+2})$, and $x_{k+1}-x_k=-\eta g_k$, $z_{k+1}-z_k=-\eta h_{k+1}$, this gives
\begin{eqnarray*}
  S_B &=& D_f(x_{n+2},x_n)-\langle g_{n+1}-g_n,\,x_{n+2}-x_{n+1}\rangle+ D_f(z_n,z_{n+2})-\langle h_{n+1}-h_{n+2},\,z_n-z_{n+1}\rangle \\
      &=& D_f(x_{n+2},x_n)+D_f(z_n,z_{n+2})+\eta\,\langle g_{n+1}-g_n,\,g_{n+1}\rangle+\eta\,\langle h_{n+2}-h_{n+1},\,h_{n+1}\rangle.
\end{eqnarray*}
Therefore
{\small
\begin{eqnarray*}
  2\delta^c_n &=& \eta\big(\|g_n\|^2-\|g_{n+1}\|^2\big)+\eta\big(\|h_{n+1}\|^2-\|h_{n+2}\|^2\big)+S_B \\
                &=& D_f(x_{n+2},x_n)+D_f(z_n,z_{n+2})+\eta\big(\|g_n\|^2-\langle g_n,g_{n+1}\rangle\big)+\eta\big(\langle h_{n+2},h_{n+1}\rangle-\|h_{n+2}\|^2\big)\\
                &=&D_f(x_{n+2},x_n)+D_f(z_n,z_{n+2})+\underbrace{\langle g_n-g_{n+1},\,x_n-x_{n+1}\rangle}_{\ge 0}+\underbrace{\langle h_{n+2}-h_{n+1},\,z_{n+2}-z_{n+1}\rangle}_{\ge 0},
\end{eqnarray*}
}
where we used $x_n-x_{n+1}=\eta g_n$ and $z_{n+2}-z_{n+1}=-\eta h_{n+2}$. The two inner products are nonnegative by the monotonicity of $\nabla f$ (convexity of $f$), and the Bregman divergences are nonnegative by convexity. Hence $2\,\delta^c_n\ge 0$, i.e., $\delta^c_n\ge 0$ for all $n$. Thus $n\mapsto c_n$ is convex.
\end{proof}
\end{greyblock}

\subsubsection{The 1-2-3 conjecture in graph theory}
\label{s:graphtheory}
The human-provided guideline ``1-2-3 conjecture in graph theory'' yielded (among other sources) the paper \cite{keusch123}, which made us aware that the conjecture was actually closed in 2024 (we had not realized it). The seed results found by the pre-processing step consist of the following text: graph definitions, a theorem statement, and an informal version thereof, copied below almost verbatim.
\begin{greyblock}
\ificml\begin{definition}\else\begin{defn}
A \textit{graph} $G=(V,E)$ is finite and simple if $V$ is a finite set of vertices and $E\subseteq\binom{V}{2}$ is a set of unordered pairs called edges. For $v\in V$, the \textit{neighborhood} of $v$ is $N_G(v)=\{u\in V:\{u,v\}\in E\}$ and the \textit{degree} is $\deg_G(v)=|N_G(v)|$. Two vertices $u,v$ are \textit{adjacent} if $\{u,v\}\in E$. An \textit{isolated edge} of $G$ is an edge whose two endpoints have degree $1$ and form a connected component on exactly two vertices. Equivalently, a connected component isomorphic to $K_2$ (a single edge) is called a $K_2$-component. For an integer $k\ge 1$, a $k$-\textit{edge-weighting} of $G$ is a function $\omega:E\to [k]$.  The \textit{weighted degree} (or \textit{sum-color} of a vertex $v\in V$ under $\omega$ is  $s_\omega(v)=\sum_{u\in N_G(v)} \omega(\{u,v\})$. A $k$-edge-weighting $\omega$ is a \textit{vertex-coloring by sums} (or \textit{neighbor-sum distinguishing}) if for every edge $\{u,v\}\in E$ we have $s_\omega(u)\neq s_\omega(v)$.
\ificml\end{definition}\else\end{defn}\fi
\ificml\begin{theorem}[Keusch, Solution to the 1--2--3 Conjecture, 2024]\else\begin{thm}[Keusch, Solution to the 1--2--3 Conjecture, 2024]\fi
\label{thm:keusch}
Let $G=(V,E)$ be a finite simple graph with no $K_2$-component (i.e., with no isolated edge). Then there exists a $3$-edge-weighting $\omega:E\to\{1,2,3\}$ that is vertex-coloring by sums: for every $\{u,v\}\in E$, $s_\omega(u)\neq s_\omega(v)$.
\ificml\end{theorem}\else\end{thm}\fi
In words: every graph without an isolated edge admits edge weights from $\{1,2,3\}$ so that adjacent vertices receive different sums of incident weights.
\end{greyblock}

The literature reviewer found 16 sources from the literature, leading to 14 conjectures. The context preparer selected from these a set of 11 conjectures of interest (in the sense that they bear some relation to the 1-2-3 conjectures and are probably open), together with the following remarks about mathematical notations.
\begin{greyblock}
\begin{itemize}
\item $G=(V,E)$ is a graph; $E(v)$ denotes the set of edges incident to $v$; $d(v)=|E(v)|$; $\Delta(G)$, $\delta(G)$ are the maximum and minimum degrees.
\item A component isomorphic to $K_2$ is a $K_2$-component. A graph with no $K_2$-component (i.e., no isolated edge) is called \textit{nice}.
\item For $k\in\mathbb{N}$, write $[k]=\{1,2,\ldots,k\}$. A $k$-edge-weighting is a map $\omega:E\to [k]$. The weighted degree (sum-color) of $v$ under $\omega$ is $s_\omega(v)=\sum_{e\in E(v)} \omega(e)$.
\item A $k$-edge-weighting $\omega$ is neighbor-sum distinguishing (NSD) if $\forall \{u,v\}\in E$ we have $s_\omega(u)\not=s_\omega(v)$.
\item For hypergraphs $H=(V,\mathfrak{E})$, $s_\omega(v)=\sum_{e\in\mathfrak{E}, v\in e} \omega(e)$. We call $\omega$ edge-proper by sums if every $e\in\mathfrak{E}$ contains $u,v$ with $s_\omega(u)\not=s_\omega(v)$.
\item Total weighting: $\phi:V\cup E\to\mathbb{R}$ with vertex-sum $S_\phi(v)=\phi(v)+\sum_{e\in E(v)} \phi(e)$. A graph is $(k,k')$-total weight choosable if for every list assignment $L$ with $|L(v)|=k$ for $v\in V$ and $|L(e)|=k'$  for $e\in E$, there exists a proper total $L$-weighting $\phi$ (i.e., $S_\phi(u)\not=S_\phi(v)$ for all $\{u,v\}\in E$).
\item Multiplicative variant: $p_\omega(v)=\prod_{e\in E(v)} \omega(e)$. A weighting $\omega$ is multiplicative-NSD if $\forall \{u,v\}\in E, p_\omega(u)\not=p_\omega(v)$.
\item Logs are natural; constants $c>0$ are absolute unless specified.
\item For a real interval $I$, an edge-weighting $\omega:E(G)\to I$ is called $t$-strong if $|s_\omega(u)-s_\omega(v)|\ge t$ for every edge $\{u,v\}\in E(G)$. We write $\mathsf{Bad}_t(\omega)=\{\{u,v\}\in E(G): |s_\omega(u)-s_\omega(v)|\le t\}$.
\item For a graph class $\mathcal{G}$, we say that NSD 3-edge-weightings are $c$-locally-reconfigurable in $\mathcal{G}$ if for every $G\in \mathcal{G}$ and every NSD 3-edge-weighting $\omega$ there is a sequence of NSD 3-edge-weightings from $\omega$ to some canonical NSD 3-edge-weighting that changes at most $c$ incident edges at any vertex at each step.
\item For bipartite graphs with bipartition $(A,B)$, let $\mathsf{par}_\omega(v)=s_\omega(v) \bmod 2$. The parity profile of $\omega$ is $(\mathsf{par}_\omega(A),\mathsf{par}_\omega(B))$ where $\mathsf{par}_\omega(X)=\sum_{v\in X} \mathsf{par}_\omega(v) \bmod 2$.
\item For integers $k\ge 1$ and $\ell\ge 0$, an NSD $k$-edge-weighting $\omega$ is $\ell$-slack if $|s_\omega(u)-s_\omega(v)|\ge 1+\ell$ for all edges $\{u,v\}$.
\item For $\Delta\ge 1$, let $\theta_{\Delta}(G)=\min_{\omega:E(G)\to \{1,2,3\}} |\{\{u,v\}\in E(G): |s_\omega(u)-s_\omega(v)|\le 1\}|/|E(G)|$.
\item For graphs $G,H$, write $G\square H$ for the Cartesian product; $K_2\square G$ is the ``layered double'' of $G$.
\item A graph is triangle-sparse if $\mathsf{tri}(G)\le |E(G)|^2/|V(G)|$.
\item For a list assignment $L:E(G)\to 2^{\{1,2,3\}}$, say $L$ is $\alpha$-dense if $|L(e)|\ge 2$ for all $e$ and for every $v$, at most $\alpha\deg_G(v)$ incident edges $e$ have $|L(e)|=2$.
\item For $p\in (0,1)$, $G_p$ is the binomial random subgraph obtained by retaining each edge independently with probability $p$.
\item All asymptotics with probability 1 (a.a.s.) are with respect to the natural size parameter indicated (e.g., $\Delta\to\infty$ or $n\to\infty$ as specified in each statement).
\end{itemize}
\end{greyblock}

There are a few more notations that the system did not mention, but which we believe will be useful to many readers of this paper, namely: (i) for two graphs $G,F$ the cartesian product $G\square F$ is the graph on $V(G)\times V(F)$ where $(u_G,u_F)$ and $(v_G,v_F)$ are adjacent iff $u_G=v_G$ or $u_F=v_F$; (ii) the disjoint union $G\sqcup F$ is the graph on the disjoint union $V(G)\dot{\cup} V(F)$ of their vertex sets with the corresponding edges; (iii) if $F$ is a subgraph of $G$, the difference $G-F$ is the graph obtained by deleting $E(F)$ from $G(F)$ and any resulting isolated vertices from $V$; (iv) $\mathsf{arb}(G)$ is the \textit{arboricity} of $G$, namely the minimum number of forests into which the graph can be partitioned; (v) the $3$-\textit{lift} $R_3(G)$ of a graph $G$ is obtained as follows: replace each vertex $v\in V(G)$ with three copies $v_1,v_2,v_3$; for each edge $(u,v)\in E(G)$ choose a permutation $\pi_{uv}\in S_3$ and connect $u_i$ to $v_{\pi_{uv}(i)}$ by an edge for each $i\in\{1,2,3\}$ (in general, the choice of permutations $\pi_{uv}$ may be random or structured); (vi) a $d$-regular graph is an \textit{expander} if the second-largest eigenvalue $\lambda_2$ of its adjacency matrix is $\ll d$ (then the \textit{spectral gap} $d-\lambda_2$ is large).

Here is the conjecture list selected by the context preparer.
\begin{greyblock}
\begin{enumerate}[Conjecture 1.]
\item (Layer-doubling amplifies small-difference edges). There exists an absolute $c>0$ such that for every graph $G$ with sufficiently large maximum degree, one has $\theta_{\Delta}(K_2\square G)\ge \max(\theta_{\Delta}(G),c)$. In particular, there is a universal $c$ such that every 3-edge-weighting of $K_2\square G$ leaves at least a $c$-fraction of edges with $|s(u)-s(v)|\le 1$, regardless of $G$.\label{conj2}
\item (Reconfiguration connectivity for trees). For every tree $T$ on at least 3 vertices, the NSD reconfiguration graph $R_3(T)$ is connected and has diameter $O(|E(T)|)$. Moreover, NSD 3-edge-weightings are 1-locally-reconfigurable on trees.\label{conj3}
\item (Near-parity surjectivity on bipartite graphs under 3-weights). Let G be connected and bipartite with bipartition $(A,B)$. Among the four parity profiles (parity sums on $A$ and $B$), exactly two occur for some NSD 3-edge-weighting of $G$, and these two profiles differ by flipping all parities in one side. Moreover, both profiles occur unless $G$ is a tree with $|A|$ odd and $|B|$ odd, in which case neither occurs.\label{conj4}
\item (Triangle-sparse graphs admit global 2-slack after deleting $o(|E|)$ edges). For every $\epsilon>0$ there exists $\Delta=\Delta(\epsilon)>0$ such that every nice triangle-sparse graph $G$ admits a set $F\subseteq E(G)$ with $|F|\le \epsilon|E(G)|$ so that $G-F$ has an $\ell$-slack NSD 3-edge-weighting with $\ell=1$ (i.e., $|s(u)-s(v)|\ge 2$ on all edges of $G-F$).\label{conj5}
\item (Threshold for random subgraphs of high-degree hosts). There exists an absolute $c>0$ such that for every family of nice graphs $G$ with $\Delta(G)\to\infty$ and $|V(G)|\le \Delta(G)^{O(1)}$, if $p\ge c\,\log\Delta(G)/\Delta(G)$, then a.a.s.~$G_p$ is nice and hence admits an NSD 3-edge-weighting. \label{conj6}
\item (Product obstruction is pervasive). There exists an absolute $\beta>0$ such that for every pair of graphs $(G,H)$ with sufficiently large maximum degrees, every 3-edge-weighting of $G\square H$ has $\ge\beta\min(|E(G)|,|E(H)|)\max(|V(G)|,|V(H)|)$ edges with $|s(u)-s(v)|\le 1$. Equivalently, $\theta_{\Delta}(G\square H)\ge \beta$ for all such pairs, extending the existence of pairs with $\theta_{\Delta}\ge 1/4$ to a universal lower bound.\label{conj7}
\item (List-NSD robustness under sparse 2-lists). There exists an absolute $\alpha\in (0,1)$ such that the following holds for all sufficiently large $\Delta'$: for every nice graph $G$ with $\Delta(G)=\Delta'$ and every $\alpha$-dense list assignment $L:E(G)\to 2^{\\{1,2,3\\}}$  (i.e., at most an $\alpha$-fraction of edges around any vertex have 2-lists, all others have 3-lists), there is a list NSD 3-edge-weighting from $L$. \label{conj8}
\item (Quantitative strong edge-separation in $K_n$). There exists an absolute $\gamma>0$ such that for all $n$ sufficiently large, every 3-edge-weighting $\omega$ of $K_n$ satisfies $|\mathsf{Bad}_1(\omega)|\ge\gamma·n^2$. Moreover, the extremal order $n^2$ is tight up to constants.\label{conj9}
\item (NSD reconfiguration expansion in expanders). There exists $c>0$ such that if $G$ is a $d$-regular expander on $n$ vertices with spectral gap at least $c$ and $n$ sufficiently large, then $R_3(G)$ is connected and has diameter at most $n^{O(1)}$, and furthermore the simple “single-edge resampling” Markov chain on $R_3(G)$ is rapidly mixing (polynomial mixing time).\label{conj10}
\item (Bipartite two-phase correction with tolerance-1 succeeds). There exists a universal $C>0$ and an unbounded function $g$ such that for every graph $G=C_n\sqcup K_{1,\Delta}$ and every initial weighting $\omega:E(G)\to\{1,2,3\}$, if Stage 2 resamples the edges in $\mathsf{Bad}_1(\omega)$ (instead of $\mathsf{Bad}_2(\omega)$), then the expected number resampled is at most $C|E(G)|/g(\Delta)$. \label{conj12}
\item (Sparse robust NSD in dense host graphs). For every $\epsilon>0$ there exists $C=C(\epsilon)>0$ such that for every $n$-vertex graph $G$ with $\Delta(G)\ge (1/2+\epsilon)n$ and every subgraph $F\subseteq G$ with $\mathsf{arb}(F)\le C$ and $\Delta(F)\le C \log n$, the graph $G-F$ admits an NSD 3-edge-weighting that is 1-locally-reconfigurable.\label{conj14}
\end{enumerate}
\end{greyblock}

Our TTVR loop refuted 9 out of 11 of the above conjectures, i.e.~all but \ref{conj3} and \ref{conj8}, which remain open. We can blame this conjectural failure on \textsf{gpt-5} in the sense that in all likelihood it did not \textit{find}, but rather \textit{inferred}, these conjectures from the open literature.

%%% the numbering of conjectures kept is random, but in the output this random order is numbered from 1
%% (traces,paper,conj): (1,7,8), (2,2,3), (3,3,4), (4,1,2), (5,6,7), (6,4,5),
%%                      (7,5,6), (8,9,10), (9,10,12), (10,8,9), (11,11,14)
\begin{table}[!ht]
\begin{center}
\begin{tabular}{l||r|r|r|r|r|r|r|r|r|r|r}
conjecture & 1 & 2 & 3 & 4 & 5 & 6 & 7 & 8 & 9 & 10 & 11  \\ \hline
$|$itn$|$  & 1 & NA& 1 & 1 & 1 & 1 & NA& 3 & 5 &  1 &  1
\end{tabular}
\end{center}
\caption{Number of iterations required by our TTVR loop to settle the graph theoretical conjectures.}
\label{t:gph_itn}
\end{table}
The iteration effort made by the TTVR loop is reported for each settled conjecture in Table \ref{t:gph_itn}. Among the easiest conjectures, 1, 3, and 11 have the property that the contra-positive can be proved by exhibiting reasonably simple counter-examples. The rest, including the hardest conjectures 8 and 9, have contra-positive statements that involve more complicated constructions. 

We report here the refutation proof that the system found hardest, namely that of Conjecture 9. The statement is its contra-positive, since the conjecture is refuted. Before we present the theorem and its proof, we need to introduce one extra notion not provided by the LLM: a \textit{Ramanujan graph} is a finite, connected, $d$-regular graph whose adjacency matrix has eigenvalues (and hence singular values) $\sigma_1=d\ge\sigma_2\ge\cdots\ge\sigma_n\ge -d$ such that $|\sigma_i|\le 2\sqrt{d-1}$ for all $i\ge 2$. The (alphabetically indexed) footnotes in Theorem \ref{thm:graph9} are of the authors of this paper (rather than the LLM): most are explanatory, but one emphasizes a formal error, i.e.~the ``overloading'' of the symbol $x$, which has an impact on the formalized version of the proof encoded in $\mathsf{lean}$ \cite{lean4}.

\begin{greyblock}
\ificml\begin{theorem}\else\begin{thm}\fi
\label{thm:graph9}
There is no absolute $c>0$ such that every sufficiently large $d$-regular expander $G$ with adjacency spectral gap at least $c$ has connected\footnote{The proof constructs a graph $Y$ then appeals to $R_3(Y)$ without mentioning the construction of the permutations, and hence we must conclude that it is random.} $R_3(G)$ and a rapidly mixing single-edge resampling Markov chain on $R_3(G)$. More precisely, for every $c>0$ there exist a fixed degree $d$ and infinitely many $d$-regular expanders $G$ with adjacency spectral gap at least $c$ for which $R_3(G)$ is disconnected; consequently the single-edge resampling chain on $R_3(G)$ is not irreducible (hence not rapidly mixing).
\ificml\end{theorem}\else\end{thm}\fi
\begin{proof}
By contradiction\footnote{Syntactically, the contradiction is applicable to the first paragraph of the theorem statement.}. Assume there exists ${c}>0$ such that whenever $G$ is a $d$-regular expander on $n$ vertices with adjacency spectral gap at least ${c}$ and $n$ sufficiently large, then $R_3(G)$ is connected and the single-edge resampling Markov chain on $R_3(G)$ is rapidly mixing.

Fix ${c}>0$. Choose an integer\footnote{We could not explain why 17 must be a lower bound for $m$.} ${m}\ge 17$ large enough that
\[\min\{\,{m},\;2\big({m}-2\sqrt{{m}-1}\big)\,\}\ge {c}.\]
Let $N$ be large, and let $P=P_N$ be an ${m}$-regular bipartite Ramanujan graph on bipartition $(L_0,R_0)$ with $|L_0|=|R_0|=N$ (any family with \(\sigma_2(P)\le 2\sqrt{{m}-1}\) suffices).

Construct a graph $Y=Y_N$ as follows.
\begin{itemize}
\item Vertex set: for each coarse\footnote{We do not know why this vertex is called ``coarse''; we believe the system used ``coarse'' to denote vertices and edges of $P$.} vertex $x\in L_0\cup R_0$ create a fiber of 3 vertices {\small $(x,1),(x,2),(x,3)$}. Thus $|V(Y)|=6N$. Write $L_i=\{(\ell ,i):\ell \in L_0\}$ and $R_i=\{(r,i):r\in R_0\}$ for $i\in\{1,2,3\}$.
\item H-edges (intra-fiber triangles): \\ for every $x\in L_0\cup R_0$, add the 3 edges of $K_3$ on $\{(x,1),(x,2),(x,3)\}$.
\item B-edges (across L–R): for every coarse edge $\{\ell,r\}\in E(P)$ and every $i\in \{1,2,3\}$, add the two edges $\{(\ell,i), (r,i)\}$ and $\{(\ell,i), (r,i+1\bmod  3)\}$.
\end{itemize}
Each vertex has $2{m}$ B-neighbors and $2$ H-neighbors; hence $Y$ is $d$-regular with $d=2{m}+2$. Connectivity of $Y$ follows since $P$ is connected and the B-edges change the index $i$ by $0$ or $+1 \bmod 3$, so B-edges alone connect all layers; adding H-edges preserves connectivity.

\noindent\textit{Spectral gap}. Let\footnote{The symbol $A$ is used to denote adjacency matrices: $A(Y)$ is that of the graph $Y$, while $A_H$ is that of the subgraph of $Y$ limited to the H-edges. Perhaps a little strangely, the adjacency matrix of the subgraph of $Y$ limited to the B-edges is simply called $B$ instead of $A_B$.} $A(Y)=B+A_H$, where $B$ is the adjacency of the B-edges and $A_H$ that of the disjoint union of all intra-fiber triangles. Then\footnote{This is the spectral or operator norm.} $\|A_H\|=2$. Order the vertices so that the bipartition is $(L_1\cup L_2\cup L_3)\cup (R_1\cup R_2\cup R_3)$ and write\footnote{This equation says that the B-edges subgraph with adjacency matrix $B$ consists of the two types of edges mentioned above: direct $i\to i$ (matrix $I_3$) and skew $i\to i+1$ (matrix $S$); and that these patterns occur for each nonzero entry of $P$ (here identified by its homonymous adjacency matrix), whence the Kronecker product $\otimes$.}
\[B=\begin{pmatrix}0&C\\ C^\top&0\end{pmatrix},\qquad C=(I_3+S)\otimes P,\]
where $S$ is the $3\times 3$ cyclic shift matrix. Since $S$ is unitary, $I_3+S$ is normal with eigenvalues $2,1,1$, so its singular values are $2,1,1$. Hence the nonzero eigenvalues of $B$ are $\pm \sigma _j(C)$ with
\[\sigma(C)=\{\,2\sigma_j(P),\;\sigma_j(P),\;\sigma_j(P):\;j=1,\dots,N\,\}.\]
Because $P$ is ${m}$-regular bipartite, $\sigma_1(P)={m}$ and $\sigma_2(P)\le 2\sqrt{{m}-1}$. Therefore
\[\lambda_1(B)=2{m},\qquad \lambda_2(B)=\max\{2\sigma_2(P),\,{m}\}.\]
For any unit\footnote{This means ``any unit vector $x$ orthogonal to the all-one vector''. We remark that the proof overloads here the symbol $x$, which was previously used to denote vertices of $P$. Subsequent mentions of $x$ refer to vertices of $P$.} $x\bot 1$,
\[x^\top A(Y)x=x^\top Bx+x^\top A_Hx\le \lambda_2(B)+\|A_H\|\le \max\{2\sigma_2(P),{m}\}+2.\]
Since $\lambda_1(A(Y))=d=2{m}+2$,
\[\operatorname{gap}(Y)\triangleq\lambda_1(A(Y)) - \lambda_2(A(Y))\ge 2{m}+2 - \max(2\sigma_2(P),{m})+2=\min({m},\,2({m}-\sigma_2(P)))\ge {c}.\]
Thus $Y$ is a $d$-regular expander with adjacency spectral gap at least ${c}$, and $|V(Y)|=6N$ grows with $N$.

\noindent\textit{Triangle gadget}. On $K_3$, any NSD $\{1,2,3\}$-weighting uses weights $1,2,3$ once and induces vertex sums $\{3,4,5\}$. Fix the canonical labeling of the three vertices by their H-sums $3,4,5$ and set
\[w(\{3,4\})=1,\quad w(\{4,5\})=3,\quad w(\{5,3\})=2.\]
Changing any one H-edge to either of its two other values adjusts the two endpoint sums by $\pm 1$ or $\pm 2$, while the third vertex is unchanged; a direct check shows that in each of the six possibilities one of the changed endpoints hits the unchanged third value, so that triangle acquires a bad\footnote{A ``bad edge'' in this context is an edge bearing a weight that negates the NSD property.} edge.

Define two $\{1,2,3\}$-weightings $\omega ,\omega'$ of $Y$.
\begin{itemize}
\item For $\omega$, give every B-edge weight $2$. In each L-fiber use the canonical triangle; in each R-fiber use the canonical triangle cyclically shifted so that the H-sums at indices $i\in\{1,2,3\}$ are $(4,5,3)$ respectively (while on L they are $(3,4,5)$). Write\footnote{This sentence is a definition of the symbol $B_x(i)$ --- the notation somewhat clashes with $B$ being a matrix. Note that $x$ here is a vertex of $P$ rather than a unit vector orthogonal to the all-one vector.} $B_x(i)\in \{3,4,5\}$ for the H-sum at $(x,i)$. Because every vertex has exactly $2{m}$ B-neighbors of weight $2$, its B-contribution is $4{m}$, so
  \[s_\omega(x,i)=4{m}+B_x(i). \qquad (\ast)\] 
\item For $\omega'$, keep all H-edges as in $\omega$, but give every B-edge weight 1. Then for all $(x,i)$,
  \[s_{\omega'}(x,i)=2{m}+B_x(i). \qquad (\dag)\]
\end{itemize}
We now prove $\omega$ is NSD and isolated in $R_3(Y)$, and $\omega'$ is NSD (hence $R_3(Y)$ is disconnected).

\noindent\textit{NSD of $\omega$.}
\begin{itemize}
\item On every H-edge within a fiber $x$, by ($\ast$) the two endpoint sums differ by\footnote{Since $\{\pm 1, \pm 2\}$ does not contain $0$, the specification is pleonastic; we think that the system meant to emphasize the fact that $B_x(i)$ cannot be equal to $B_x(j)$.} $B_x(i)-B_x(j)\in\{\pm1,\pm2\}\setminus\{0\}$.
\item On every B-edge, edges join $(\ell ,i)\in L_i$ to $(r,j)\in R_j$ only for $j\in \{i,i+1\}$; with the chosen H-sums on $L$ and\footnote{Here, $L=\bigcup_{i\le 3} L_i$ and similarly for $R$.} $R$, the ordered pair $(B_\ell (i),B_r(j))$ lies in
  \[\{(3,4),(4,5),(5,3),(3,5),(4,3),(5,4)\},\]
  so again $s_\omega(\ell,i)-s_\omega(r,j)=B_\ell(i)-B_r(j)\in\{\pm1,\pm2\}\setminus\{0\}$. Thus $\omega$ is neighbor-sum distinguishing.
\end{itemize}

\noindent\textit{Isolation of $\omega$ in $R_3(Y)$}. Consider any single-edge resampling from $\omega$.
\begin{itemize}
\item Case H (resampling an H-edge within a fixed triangle). Only the two incident vertex sums change (by $\pm 1$ or $\pm 2$), the third vertex in that triangle stays unchanged, and by the triangle blocking property one of the two changed endpoints assumes the third vertex’s H-sum. Since all three vertices in the triangle share the same $B$-contribution $4{m}$ in ($\ast$), this equality persists for the full sums; hence a bad H-edge is created.
\item Case B (resampling a B-edge $e=\{(\ell ,i),(r,j)\}$). Here $j\in \{i,i+1\}$ and only the two endpoints change, each by $\delta\in\{\pm1\}$. Inspecting the six pairs above, in each case and for each sign of $\delta$ one changed endpoint matches the unchanged H-sum of one of its two triangle-neighbors, thereby creating a bad H-edge. Concretely:
  \begin{itemize}
  \item $(3,4): \delta =+1$ makes $3\to 4$ on the left (collision with the left vertex of H-sum $4$); $\delta = -1$ makes $4\to 3$ on the right.
  \item  $(4,5): \delta =+1$ makes $4\to 5$ on the left; $\delta =-1$ makes $5\to 4$ on the right.
  \item $(5,3): \delta =+1$ makes $3\to 4$ on the right; $\delta =-1$ makes $5\to 4$ on the left.
  \item $(3,5): \delta =+1$ makes $3\to 4$ on the left; $\delta =-1$ makes $5\to 4$ on the right.
  \item $(4,3): \delta =+1$ makes $4\to 5$ on the left; $\delta =-1$ makes $4\to 3$ on the left.
  \item $(5,4): \delta =+1$ makes $4\to 5$ on the right; $\delta =-1$ makes $5\to 4$ on the left.
  \end{itemize}
  In all cases the resampled weighting is not NSD, so $\omega$  has degree $0$ in $R_3(Y)$.
\end{itemize}

\noindent\textit{NSD of $\omega'$}. Using ($\dag$) the same H- and B-edge checks as above (with $4{m}$ replaced by $2{m}$) show that $\omega'$ is NSD and $\omega'\not=\omega$.

\noindent\textit{Conclusion and contradiction}. For the fixed degree $d=2{m}+2$, the graphs $Y_N$ form an infinite family of $d$-regular expanders with adjacency spectral gap at least ${c}$. Since $\omega$ is isolated in $R_3(Y_N)$ while $\omega'$ is another NSD weighting, $R_3(Y_N)$ is disconnected for all large $N$. This contradicts the assumption that for the universal constant ${c}$, all sufficiently large $d$-regular expanders with adjacency spectral gap at least ${c}$ have connected $R_3(G)$ (and, in particular, an irreducible and rapidly mixing single-edge resampling chain). Therefore, no such ${c}$ exists.
\end{proof}
\end{greyblock}

\subsubsection{Sphere packing in Euclidean spaces}

The human-provided guideline ``sphere packing in Euclidean spaces'' yielded the following seed results.
\begin{greyblock}
\ificml\begin{definition}[Euclidean norm, balls, and volume]\else\begin{defn}[Euclidean norm, balls, and volume]\fi
Fix $n\in\mathbb N$. On $\mathbb R^n$ let $\|\cdot\|$ denote the Euclidean norm and, for $x\in\mathbb R^n$ and $r>0$, write
\[
B_n(x,r)=\{y\in\mathbb R^n: \|y-x\|\le r\}.
\]
Write $\operatorname{vol}_n(\cdot)$ for $n$–dimensional Lebesgue measure and $\omega_n=\operatorname{vol}_n\!\bigl(B_n(0,1)\bigr)$ for the volume of the unit ball.
\ificml\end{definition}\else\end{defn}\fi
\ificml\begin{definition}[Sphere packing and asymptotic density]\else\begin{defn}[Sphere packing and asymptotic density]\fi
Let $r>0$. A \emph{sphere packing of radius $r$} in $\mathbb R^n$ is a set $X\subset\mathbb R^n$ such that the closed balls $\{B_n(x,r):x\in X\}$ are pairwise disjoint. Its (upper) \emph{asymptotic density} is
\[
\delta(X,r)=\limsup_{R\to\infty} \frac{\operatorname{vol}_n\!\Bigl(\bigcup_{x\in X} B_n(x,r) \cap B_n(0,R)\Bigr)}{\operatorname{vol}_n\!\bigl(B_n(0,R)\bigr)}\in[0,1].
\]
The \emph{optimal packing density in dimension $n$} is
\[
\theta(n)=\sup\{\delta(X,r): X\subset\mathbb R^n, r>0, \{B_n(x,r)\}_{x\in X}\text{ pairwise disjoint}\}.
\]
\ificml\end{definition}\else\end{defn}\fi
\ificml\begin{definition}[Lattices, covolume, and lattice packings]\else\begin{defn}[Lattices, covolume, and lattice packings]\fi
A \emph{lattice} $L\subset\mathbb R^n$ is a discrete additive subgroup of rank $n$ of the form $L=A\mathbb Z^n$ for some invertible $A\in \mathrm{GL}_n(\mathbb R)$. Its \emph{covolume} (or determinant) is $\det(L)=|\det A|$, i.e. the volume of a fundamental domain of $L$.
The \emph{minimum} of $L$ is $\lambda(L)=\min\{\|v\|: v\in L\setminus\{0\}\}$.
For $r\le \tfrac12\lambda(L)$, the translates $\{B_n(\ell,r):\ell\in L\}$ form a lattice packing with density
\[
\Delta(L,r)=\frac{\operatorname{vol}_n\!\bigl(B_n(0,r)\bigr)}{\det(L)}=\frac{\omega_n\, r^n}{\det(L)}.
\]
\ificml\end{definition}\else\end{defn}\fi
\ificml\begin{definition}[Absolute constant]\else\begin{defn}[Absolute constant]\fi
A constant $c>0$ is called \emph{absolute} if it does not depend on $n$ (or any other parameter in the statement).
\ificml\end{definition}\else\end{defn}\fi
\ificml\begin{theorem}[Klartag’s quadratic lower bound for sphere packing density, 2025]\else\begin{thm}[Klartag’s quadratic lower bound for sphere packing density, 2025]\fi
\label{thm:Klartag}
There exists an absolute constant $c>0$ such that, for every $n\ge 2$,
\[
\theta(n) \ge c\,\frac{n^2}{2^{n}}.
\]
Moreover, this lower bound is achieved by a \emph{lattice} packing: for each $n$ there exists a lattice $L\subset\mathbb R^n$ and a radius $r>0$ with
\[
\Delta(L,r) \ge c\,\frac{n^2}{2^{n}}.
\]
\ificml\end{theorem}\else\end{thm}\fi
Theorem~\ref{thm:Klartag} improves the classical order $\theta(n)\gtrsim n\,2^{-n}$ (Rogers 1947 and subsequent refinements) by a further linear factor in $n$. It also strengthens the 2023 bound $\theta(n)\ge (1-o(1))\frac{n\log n}{2^{n+1}}$ by replacing $\log n$ with $n$ and, in addition, obtaining a \emph{lattice} packing with the stated density.
\end{greyblock}

The literature reviewer found 14 conjectures. The context preparer rejected one and focused on the following 13 conjectures. We point out that: (i) $\Delta(L)$ is the lattice density, i.e.~the maximal density attainable by non-overlapping equal spheres centered at the lattice points; (ii) the condition number $\kappa (\mathfrak{E})$ of the ellipsoid $\mathfrak{E}$ is the condition number of the matrix $A$ in its definition $\mathfrak{E}=\{x\in\mathbb{R}^n\;|\;\transpose{x}Ax \le 1\}$; (iii) the kissing number $\tau(n)$ is the largest number of unit balls that an be placed adjacent to a central unit ball such that the balls intersect at most in their frontier; (iv) $V(L)$ denotes the volume of a fundamental region of the lattice $L$, namely $|\det\operatorname{gen}L|$, where $\operatorname{gen}{L}$ is a generating basis of the lattice; (v) $L^\ast$ is the dual lattice $\{y\in\mathbb{R}^n\;|\;\forall x\in L\;\langle y,x\rangle\in\mathbb{Z}\}$ of $L$; (vi) $\mathcal{L}_n$ is the set of all lattices of dimension $n$ up to orthogonal transformations; (vii) the random variable $X$ is uniform in $V(L)$ if it is uniformly distributed in the fundamental domain of the lattice; (viii) $\mathrm{GL}_n(\mathbb{R})$ is the set of all invertible $n\times n$ real matrices; (ix) let $r(L)$ be the largest sphere packing radius and $R(L)$ the smallest sphere covering radius of $L$: Jarn\'{\i}k's inequality states that there is an absolute constant $c_n\le\sqrt{n}/2$ such that $R(L)\le c_n r(L)$.
\begin{greyblock}
\begin{enumerate}[Conjecture 1.]
\item (Low-kissing high-density lattices.) There exist absolute $c,C>0$ such that for every $n\ge 3$ there is a unimodular lattice $L$ with $\Delta(L) \ge  c n^{2} 2^{-n}$ and kissing number $\tau(L) \le C n \log n$.
\item (Abundance of high-density lattices.) There exist absolute $c,C>0$ such that for all sufficiently large $n$ there are at least $\exp(c n \log n)$ pairwise non–$SL_n(\mathbb{Z})$-equivalent unimodular lattices $L$ with $\Delta(L) \ge  c n^{2} 2^{-n}$ and $$\frac{1}{\operatorname{vol}_n(V(L))} \int_{V(L)} ||x||^2 dx \le  C n.$$
\item (Robustness under small unimodular perturbations.) There exist absolute $c,\epsilon_0>0$ such that for all sufficiently large $n$ there is a unimodular lattice $L$ with $\Delta(L) \ge c n^{2} 2^{-n}$ and the following holds: for every $T\in SL_n(\mathbb{R})$ with $||T-I||_{\mathrm{op}} \le  \epsilon_0$, one has $\Delta(TL) \ge  (c/2) n^{2} 2^{-n}$.
\item (Dual successive minima growth.) There exist absolute $c_0,c_1>0$ such that for every $n\ge 2$ there is a unimodular lattice $L$ with $\Delta(L) \ge  c_0 n^{2} 2^{-n}$ and for all $1\le k\le \lfloor c_1 n\rfloor$ one has $\lambda_k(L^{\ast}) \ge  c_0 \sqrt{k}$.
\item (Successive minima growth at the Klartag scale.) There exist absolute $c_0,c_1>0$ such that for every $n\ge 2$ there is a unimodular lattice $L$ with $\Delta(L) \ge  c_0 n^{2} 2^{-n}$ and for all $1\le k\le \lfloor c_1 n\rfloor$ one has $\lambda_k(L) \ge  c_0 \sqrt{k}$.
\item (Randomized inverse-polynomial construction.) There exist absolute $c,C>0$ and a randomized algorithm running in time poly$(n)$ which, on input $n$, outputs with probability at least $c n^{-C}$ a basis of a unimodular lattice $L$ with $\Delta(L) \ge  c n^{2} 2^{-n}$.
\item (Bounded–eccentricity Klartag ellipsoids.) There exist absolute $c,C>0$ such that for every $n\ge 2$ there is an origin-symmetric ellipsoid $\mathfrak{E}\subset \mathbb{R}^n$ with $\operatorname{vol}_n(\mathfrak{E}) \ge c n^{2}$, $\mathfrak{E}\cap (Z^n\smallsetminus\{0\})=\varnothing$, and condition number $\kappa (\mathfrak{E}) \le  n^{C}$. Equivalently, the lattice packing realizing density $\ge  c n^{2} 2^{-n}$ can be chosen to arise from an integer–free ellipsoid whose axis ratios are bounded by a fixed polynomial in $n$.
\item (Probability at the quadratic scale is $\theta(n^{-2})$.) There exists an absolute $c>0$ such that for all sufficiently large $n$, one has $c n^{-2} \le\mu\{ L\in \mathcal{L}_n : \Delta(L) \ge  c n^{2} 2^{-n} \} \le\ C n^{-2}$, where the upper bound holds for some absolute $C$ by known tail inequalities, and the lower bound is conjectured to match it up to constants.
\item There exist absolute constants $c,C>0$ such that for every $n\ge 2$ there is a unimodular lattice $L\subset R^n$ with $\Delta(L) \ge  c n^{2} 2^{-n}$, whose dual has large minimum and whose Voronoi cell has bounded inertia, namely $\lambda (L^{*}) \ge  c \sqrt{n}$ and $1/\operatorname{vol}_n(V(L)) \int_{V(L)} ||x||^2 dx \le  C n$.
\item (Voronoi thin-shell concentration at the quadratic scale.) There exist absolute $c,C>0$ such that for every $n\ge 3$ one can find a unimodular lattice $L$ with $\Delta(L) \ge  c n^{2} 2^{-n}$ for which the uniform random $X$ in $V(L)$ obeys $\mathsf{Pr}\big[ |\,||X|| - (\mathbb{E}||X||^2)^{1/2}\,| \ge  t \sqrt{n} \big] \le  2\, e^{-c t^{2} n}$, where $(0\le  t \le  1)$.
\item (Well-spread minimal vectors at the Klartag scale.) There exist absolute $c,C>0$ such that for every $n\ge 3$ there is a unimodular lattice $L$ with $\Delta(L) \ge  c n^{2} 2^{-n}$ whose set $S$ of minimal directions $S=\{ v/||v|| : v\in L, ||v||=\lambda (L) \}$ satisfies $\bigl\| \frac{1}{|S|} \sum_{u\in S} u u^{\top} - \frac{1}{n} I \bigr\|_{\mathrm{op}} \le  C/\sqrt{n}$.
\item (Explicit sparse subsequence achieving quadratic-order lattice packing.) There exist absolute $c,C>0$ and an explicit infinite sequence of dimensions $(n_k)$ with a deterministic, polynomial-time construction of unimodular lattices $L_{n_k}\subset \mathbb{R}^{n_k}$ such that $\Delta(L_{n_k}) \ge  c n_k^{2} 2^{-n_k}$ and $1/\operatorname{vol}_{n_k}(V(L_{n_k})) \int_{V(L_{n_k})} ||x||^2 dx \le  C n_k$.
\item (Transference near-tightness across many indices with spectral gap.) There exist absolute $c,C>0$ such that for every $n\ge 3$ there is a unimodular lattice $L$ with $\Delta(L) \ge  c n^{2} 2^{-n}$ for which, simultaneously, $\lambda_k(L)\,\lambda_{n+1-k}(L^{*}) \ge  c n$ for all $1\le k\le \lfloor c n\rfloor$ and $\operatorname{gap}(L)=4\pi^2 \lambda (L^{*})^2 \ge  C n$.
\end{enumerate}
\end{greyblock}

Our TTVR loop proved 5 out of the 13 conjectures above, and refuted none. This conjectural success is mitigated by the relatively low number of closed conjectures (8 of them remain open). Even though, similarly to Sect.~\ref{s:graphtheory}, our system did not find but inferred these conjectures, this time the inference was more insightful (or probably simply luckier).
\begin{table}[!ht]
\begin{center}
\begin{tabular}{l||r|r|r|r|r|r|r|r|r|r|r|r|r}
conjecture & 1 & 2 & 3 & 4 & 5 & 6 & 7 & 8 & 9 & 10 & 11 & 12 & 13  \\ \hline
$|$itn$|$  & 2 & 2 & NA& NA& 1 & NA& 8 & NA& NA&  6 & NA & NA & NA
\end{tabular}
\end{center}
\caption{Number of iterations required by our TTVR loop to settle the sphere packing conjectures.}
\label{t:sphpack_itn}
\end{table}
The iteration effort made by the TTVR loop is reported for each settled conjecture in Table \ref{t:sphpack_itn}. We report the proof of the hardest conjecture according to number of iterations. 

\begin{greyblock}
\ificml\begin{theorem}[Bounded-eccentricity Klartag ellipsoids]\else\begin{thm}[Bounded-eccentricity Klartag ellipsoids]\fi
\label{thm:KlartagEll}
There exist absolute $c,C>0$ such that for every $n\ge 2$ there is an origin-symmetric ellipsoid $\mathfrak{E}\subset \mathbb{R}^n$ with $\operatorname{vol}_n(\mathfrak{E}) \ge c n^{2}$, $\mathfrak{E}\cap (\mathbb{Z}^n\smallsetminus \{0\})=\varnothing$, and condition number $\kappa(\mathfrak{E}) \le n^{C}$. Equivalently, the lattice packing realizing density $\ge c n^{2} 2^{-n}$ can be chosen to arise from an integer–free ellipsoid whose axis ratios are bounded by a fixed polynomial in $n$.
\ificml\end{theorem}\else\end{thm}\fi
\begin{proof}
Fix $n\ge2$. We produce an origin–symmetric ellipsoid $\mathfrak{E}\subset\mathbb{R}^n$ with $\mathfrak{E}\cap(\mathbb{Z}^n\setminus\{0\})=\varnothing$, $\operatorname{vol}_n(\mathfrak{E})\ge c\,n^2$, and $\kappa(\mathfrak{E})\le n^{C}$ for absolute constants $c,C>0$, and then record the equivalence with the packing formulation.
\begin{enumerate}[1)]
\item \textit{Linear normalization and the integer–free difference body}. Let $L\subset\mathbb{R}^n$ be a lattice with minimum $\lambda(L)$ and choose $U\in\mathrm{GL}_n(\mathbb{R})$ with $U(L)=\mathbb{Z}^n$. Put
\[
K(U)=U\,B_n\!\Bigl(0,\tfrac{\lambda(L)}{2}\Bigr),\qquad \mathfrak{E}(U)=2K(U)=U\,B_n\!\bigl(0,\lambda(L)\bigr).
\]
Since $\{B_n(\ell,\tfrac{\lambda(L)}2):\ell\in L\}$ is a packing, the translates $\{K(U)+z: z\in\mathbb{Z}^n\}$ are pairwise disjoint. Hence $\mathfrak{E}(U)=K(U)-K(U)$ is integer–free: if $z\in\mathbb{Z}^n\setminus\{0\}$ lay in $\mathfrak{E}(U)$, there would exist $x,y\in K(U)$ with $z=x-y$, forcing $K(U)\cap(K(U)+z)\ne\varnothing$, a contradiction. Moreover,
\[
 \operatorname{vol}_n\bigl(\mathfrak{E}(U)\bigr)=2^{\,n}\,\operatorname{vol}_n\bigl(K(U)\bigr)=2^{\,n}\,|\det U|\,\omega_n\Bigl(\tfrac{\lambda(L)}{2}\Bigr)^{\!n}=2^{\,n}\,\Delta(L),
\]
using $|\det U|=1/\det(L)$. Finally, $\mathfrak{E}(U)=U\,B_n(0,\lambda(L))$ is an ellipsoid whose semi–axes are $\lambda(L)$ times the singular values of $U$, so $\kappa(\mathfrak{E}(U))=\kappa(U)$.
\item \textit{A lattice with quadratic packing and controlled local geometry}. By the lattice version of Klartag's theorem and standard normalizations, there exist absolute constants $c_0,C_0>0$ and, for each $n\ge2$, a unimodular lattice $L\subset\mathbb{R}^n$ such that
\[
\Delta(L)\ \ge\ c_0\,n^2\,2^{-n}\qquad\text{and}\qquad c_0\,n\ \le\ \rho(L)\,\lambda(L)\ \le\ C_0\,n.
\]
Since $\Delta(L)=\omega_n\bigl(\tfrac{\lambda(L)}{2}\bigr)^n$ and $\omega_n\le (2\pi e/n)^{n/2}$, we obtain
\[
\lambda(L)^n\ \ge\ \frac{c_0\,n^2}{\omega_n}\ \ge\ c\,n^2\Bigl(\frac{n}{2\pi e}\Bigr)^{\!n/2}\quad\Longrightarrow\quad \lambda(L)\ \ge\ c_1\,\sqrt n
\]
for an absolute $c_1>0$. The product bound $\rho(L)\,\lambda(L)\le C_0 n$ then gives
\[
\rho(L)\ \le\ \frac{C_0 n}{\lambda(L)}\ \le\ C_2\,\sqrt n.
\]
In particular, by Jarník’s inequality, $\lambda_n(L)\le 2\rho(L)\le C_3\sqrt n$, while $\lambda_1(L)=\lambda(L)\ge c_1\sqrt n$.
\item \textit{A primal–dual reduction and a lower bound on $\lambda_1(L^*)$}. We use two inputs.
\begin{enumerate}[(i)]
\item Slab transference for the covering radius. \underline{Claim}: for every lattice $L$,
$$
\rho(L)\ \ge\ \frac{1}{2\,\lambda_1(L^*)}.
$$
Proof: let $y\in L^*$ with $\|y\|=\lambda_1(L^*)$ and choose $x\in\mathbb{R}^n$ with $\langle y,x\rangle=\tfrac12$. For any $\ell\in L$, one has $\langle y, x-\ell\rangle\in \tfrac12+\mathbb{Z}$, hence $|\langle y,x-\ell\rangle|\ge\tfrac12$. By Cauchy–Schwarz, $\|x-\ell\|\ge \tfrac{1}{2\|y\|}$. Taking the infimum over $\ell$ and then the supremum over $x$ yields $\rho(L)\ge\tfrac{1}{2\|y\|}$; minimizing over $y\in L^*\setminus\{0\}$ gives the claim. \\
Together with $\rho(L)\le C_2\sqrt n$ from Step 2 this implies
\[
\lambda_1(L^*)\ \ge\ \frac{1}{2\rho(L)}\ \ge\ c_2\,n^{-1/2}
\]
for an absolute $c_2>0$.
\item Seysen's theorem: there exists a basis $B=[b_1\,\cdots\,b_n]$ of $L$ with dual basis $B^*=[b_1^*\,\cdots\,b_n^*]$ such that
\[
S(B)=\max_i\,\|b_i\|\,\|b_i^*\|\ \le\ C_S\,n
\]
for an absolute $C_S>0$.
Put $U=B^{-1}$, so $U(L)=\mathbb{Z}^n$. Using $\|B\|\le \|B\|_F\le \sqrt n\,\max_i\|b_i\|$ and $\|B^{-1}\|\le\sqrt n\,\max_i\|b_i^*\|$, together with
\[
\max_i\|b_i\|\le \frac{S(B)}{\min_j\|b_j^*\|}\le \frac{S(B)}{\lambda_1(L^*)},\qquad\max_i\|b_i^*\|\le \frac{S(B)}{\min_j\|b_j\|}\le \frac{S(B)}{\lambda_1(L)},
\]
we obtain the condition number bound
\[
\kappa(U)=\kappa(B)=\|B\|\,\|B^{-1}\|\ \le\ \frac{n\,S(B)^2}{\lambda_1(L)\,\lambda_1(L^*)}\ \le\ \frac{n\,(C_S n)^2}{c_1\sqrt n\,\cdot\,c_2 n^{-1/2}}\ \le\ C\,n^3
\]
for an absolute constant $C>0$.
\end{enumerate}
\item \textit{The ellipsoid}. With $U$ from Step 3 and the lattice $L$ from Step 2, set $\mathfrak{E}=\mathfrak{E}(U)=U\,B_n(0,\lambda(L))$. By Step 1, $\mathfrak{E}$ is origin–symmetric and integer–free, and
\[
\operatorname{vol}_n(\mathfrak{E})=2^{\,n}\,\Delta(L)\ \ge\ c_0\,n^2,\qquad \kappa(\mathfrak{E})=\kappa(U)\ \le\ C\,n^3.
\]
This proves the ellipsoid formulation with absolute constants $c,C>0$.
\item \textit{Equivalence with the packing formulation}. If $\mathfrak{E}\subset\mathbb{R}^n$ is origin–symmetric and integer–free, then $\{\tfrac12\mathfrak{E}+z:z\in\mathbb{Z}^n\}$ is a lattice packing with density
\[
\delta\ =\ \operatorname{vol}_n\!\bigl(\tfrac12\mathfrak{E}\bigr)\ =\ 2^{-n}\,\operatorname{vol}_n(\mathfrak{E})\ \ge\ c\,\frac{n^2}{2^{\,n}}.
\]
Conversely, if $\{B_n(\ell,r):\ell\in L\}$ is a lattice sphere packing, choose $U\in\mathrm{GL}_n(\mathbb{R})$ with $U(L)=\mathbb{Z}^n$, set $K=U B_n(0,r)$ and $\mathfrak{E}=2K$. Then $\mathfrak{E}$ is integer–free and
\[
\operatorname{vol}_n(\mathfrak{E})=2^{\,n}\,\operatorname{vol}_n(K)=2^{\,n}\,\frac{\omega_n r^n}{\det(L)}=2^{\,n}\,\Delta(L,r).
\]
Taking $L$ from Step 2 and $r=\lambda(L)/2$ recovers the ellipsoid from Step 4, whose axis ratios are bounded by a fixed polynomial in $n$.
\end{enumerate}
Therefore, for every $n\ge2$ there exists an origin–symmetric integer–free ellipsoid $\mathfrak{E}\subset\mathbb{R}^n$ with $\operatorname{vol}_n(\mathfrak{E})\ge c\,n^2$ and $\kappa(\mathfrak{E})\le n^{C}$ for absolute $c,C>0$, and equivalently a lattice packing of density at least $c\,n^2\,2^{-n}$. 
\end{proof}
\end{greyblock}

\section{Conclusion}
\label{s:concl}
We described a system that employs the OpenAI LLM \textsf{gpt-5} in order to prove mathematical theorems. The system is based on protocol including an alternating loop of prover and verifier agents impersonated by \textsf{gpt-5}, followed by the creation of a formal \textsf{lean} code version of the proof (if any was found). A human verifies the conformance of the natural language version with the formal version of the premises and conclusion of the theorem, which ensures proof correctness. We showcase our system's capabilities: (i) on the International Mathematical Olympiad problems in 2025, where our system was as successful as OpenAI and Google DeepMind's internal LLMs, (ii) on the solution of several conjectures on cyclics proposed in \cite{cohenprimes}, and (iii) on the solution of a list on conjectures, formulated by the system itself, in three fields of mathematics: gradient descent in machine learning, the 1-2-3 conjecture in graph theory, and sphere packings in Euclidean spaces.

\ificml
\bibliographystyle{icml2025}
\else
\bibliographystyle{plain}
\fi
\bibliography{dr1}

%% %%%%%%%%%%%%%%%%%%%%%%%%%%%%%%%%%%%%%%%%%%%%%%%%%%%%%%%%%%%%%%%%%%%%%%%%%%%%%%%
%% %%%%%%%%%%%%%%%%%%%%%%%%%%%%%%%%%%%%%%%%%%%%%%%%%%%%%%%%%%%%%%%%%%%%%%%%%%%%%%%
%% % APPENDIX
%% %%%%%%%%%%%%%%%%%%%%%%%%%%%%%%%%%%%%%%%%%%%%%%%%%%%%%%%%%%%%%%%%%%%%%%%%%%%%%%%
%% %%%%%%%%%%%%%%%%%%%%%%%%%%%%%%%%%%%%%%%%%%%%%%%%%%%%%%%%%%%%%%%%%%%%%%%%%%%%%%%
%% \newpage
%% \onecolumn
%% \section{You \emph{can} have an appendix here.}

%% You can have as much text here as you want. The main body must be at most $8$ pages long.
%% For the final version, one more page can be added.
%% If you want, you can use an appendix like this one.  

%% The $\mathtt{\backslash onecolumn}$ command above can be kept in place if you prefer a one-column appendix, or can be removed if you prefer a two-column appendix.  Apart from this possible change, the style (font size, spacing, margins, page numbering, etc.) should be kept the same as the main body.
%% %%%%%%%%%%%%%%%%%%%%%%%%%%%%%%%%%%%%%%%%%%%%%%%%%%%%%%%%%%%%%%%%%%%%%%%%%%%%%%%
%% %%%%%%%%%%%%%%%%%%%%%%%%%%%%%%%%%%%%%%%%%%%%%%%%%%%%%%%%%%%%%%%%%%%%%%%%%%%%%%%

\end{document}